%% file: main_cr.tex
\documentclass[10pt,twocolumn,letterpaper]{article}

\usepackage{cvpr}              
\usepackage[accsupp]{axessibility}

\input{preamble}

\definecolor{cvprblue}{rgb}{0.21,0.49,0.74}
\usepackage[pagebackref,breaklinks,colorlinks,allcolors=cvprblue]{hyperref}

\def\paperID{} 
\def\confName{CVPR}
\def\confYear{2026}

\title{PBSBench: A Multi-Level Vision-Language Framework and Benchmark for Hematopathology Whole Slide Image Interpretation}

\author{
  Yuanlong Wang$^1$ \quad
  Weichi Chen$^1$ \quad
  Adrian Rajab$^2$ \quad
  Wenfang Liu$^2$ \quad \\
  Yulan Jin$^2$ \quad
  Andrew Srisuwananukorn$^2$$^\dagger$ \quad   
  Ping Zhang$^1$$^\dagger$ \\    
  $^1$The Ohio State University \quad $^2$The Ohio State University Wexner Medical Center \\
  {\tt\small \{wang.16050, chen.12915, zhang.10631\}@osu.edu}\\
  {\tt\small \{Adrian.Rajab, Wenfang.Liu, Yulan.Jin, sris01\}@osumc.edu}
}

\begin{document}
\maketitle

\renewcommand{\thefootnote}{\fnsymbol{footnote}}
\footnotetext[2]{Co-corresponding authors.}
\renewcommand{\thefootnote}{\arabic{footnote}}
\input{sec_cr/0_abstract}    
\input{sec_cr/1_intro}

\input{sec_cr/2_related_works}
\input{sec_cr/3_method}

\input{sec_cr/4_experiments}
\input{sec_cr/5_conclusion}
\section*{Acknowledgement}
This work was funded in part by the National Science Foundation under award number IIS-2145625 and by the National Institutes of Health under awards number R01AI188576 and R01CA301579.

{
    \small
    \bibliographystyle{ieeenat_fullname}
    \bibliography{main}
}

\input{sec_cr/A_data_stats}
\input{sec_cr/B_ethical_statement}
\input{sec_cr/C_data_processing}
\input{sec_cr/D_prompts}
\input{sec_cr/E_additional_performance}

\input{sec_cr/F_case_study}

\input{sec_cr/G_large_tabels}

\end{document}

%% file: preamble.tex
\usepackage{multirow}



%% file: sec_cr/0_abstract.tex
\begin{abstract}
Peripheral Blood Smear (PBS) is a critical microscopic examination in hematopathology that yields whole-slide imaging (WSI). Unlike solid tissue pathology, PBS interpretation focuses on individual cell morphologies rather than tissue architecture, making it distinct in both visual characteristics and diagnostic reasoning. However, current  multimodal large language models (MLLMs) for pathology are primarily developed on solid-tissue WSIs and struggle to generalize to PBS. To bridge this gap, we construct \textbf{PBSInstr}, the first vision-language dataset for PBS interpretation, comprising 353 PBS WSIs paired with microscopic impression paragraphs and 29k cell-level image crops annotated with cell type labels and morphological descriptions. To facilitate instruction tuning, PBSInstr further includes 27k question-answer (QA) pairs for cell crops and 1,286 QA pairs for PBS slides. Building upon PBSInstr, we develop \textbf{PBS-VL}, a hematopathology-tailored vision-language model for multi-level PBS interpretation at both cell and slide levels. To comprehensively evaluate PBS understanding, we construct \textbf{PBSBench}, a visual question answering (VQA) benchmark featuring four question categories and six PBS interpretation tasks. Experiments show that PBS-VL outperforms existing general-purpose and pathology MLLMs, underscoring the value of PBS-specific data. We release our code\footnote{https://github.com/Wang-Yuanlong/PBSBench}, datasets, and model weights to facilitate future research. Our proposed framework lays the foundation for developing practical AI assistants supporting decision-making in hematopathology.
\end{abstract}

%% file: sec_cr/1_intro.tex
\begin{figure}
    \centering
    \begin{subfigure}{0.48\linewidth}
    \includegraphics[width=\textwidth]{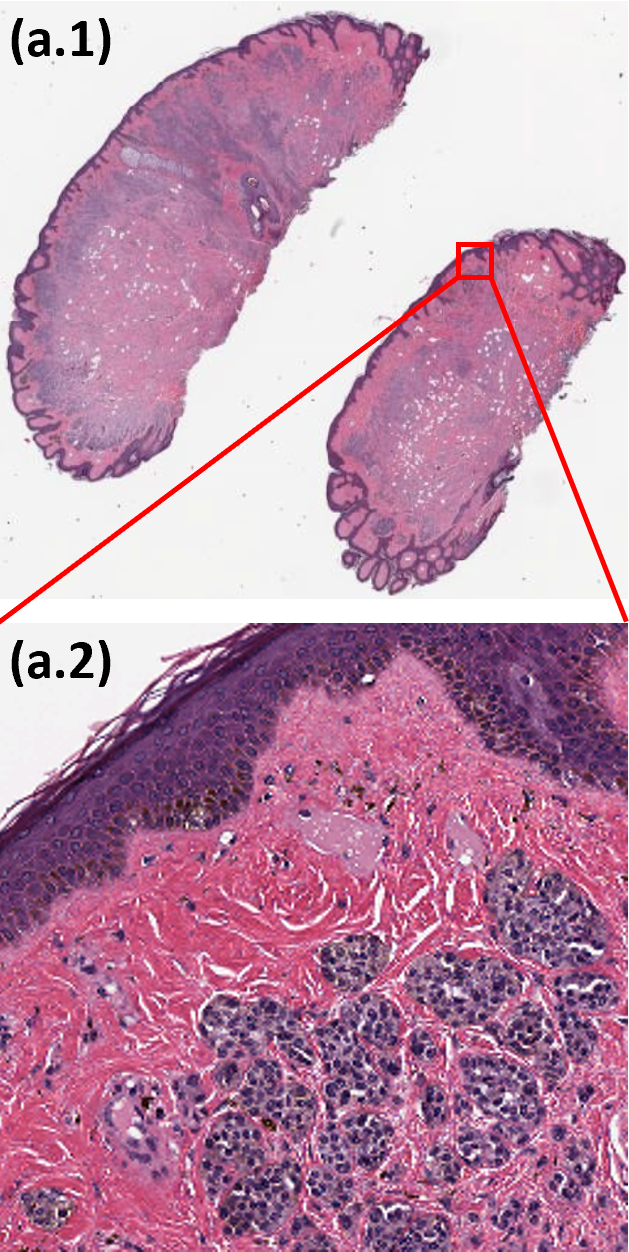}
    \caption{Histopathology Slides}
    \label{fig:comparison-a}
    \end{subfigure}
      \hfill
    \begin{subfigure}{0.48\linewidth}
    \includegraphics[width=\textwidth]{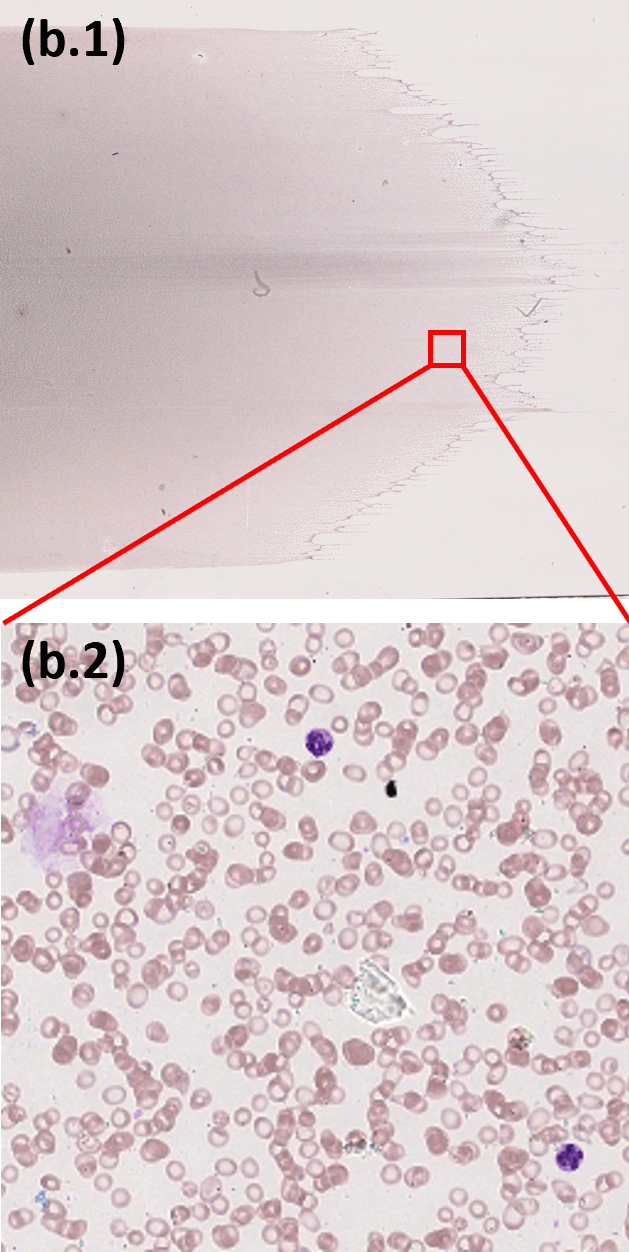}
    \caption{PBS slides}
    \label{fig:comparison-b}
  \end{subfigure}
    \caption{Visual comparison between histopathology whole slides and peripheral blood smears. \cref{fig:comparison-a} Histopathology whole slides have a clear tissue boundary and structure; \cref{fig:comparison-b} PBS spreads across the slide without tissue structure.}
    \label{fig:wsi-comparison}
\end{figure}

\section{Introduction}
\label{sec:intro}

\begin{figure*}
  \centering
  \begin{subfigure}{0.46\linewidth}
    \includegraphics[width=\textwidth]{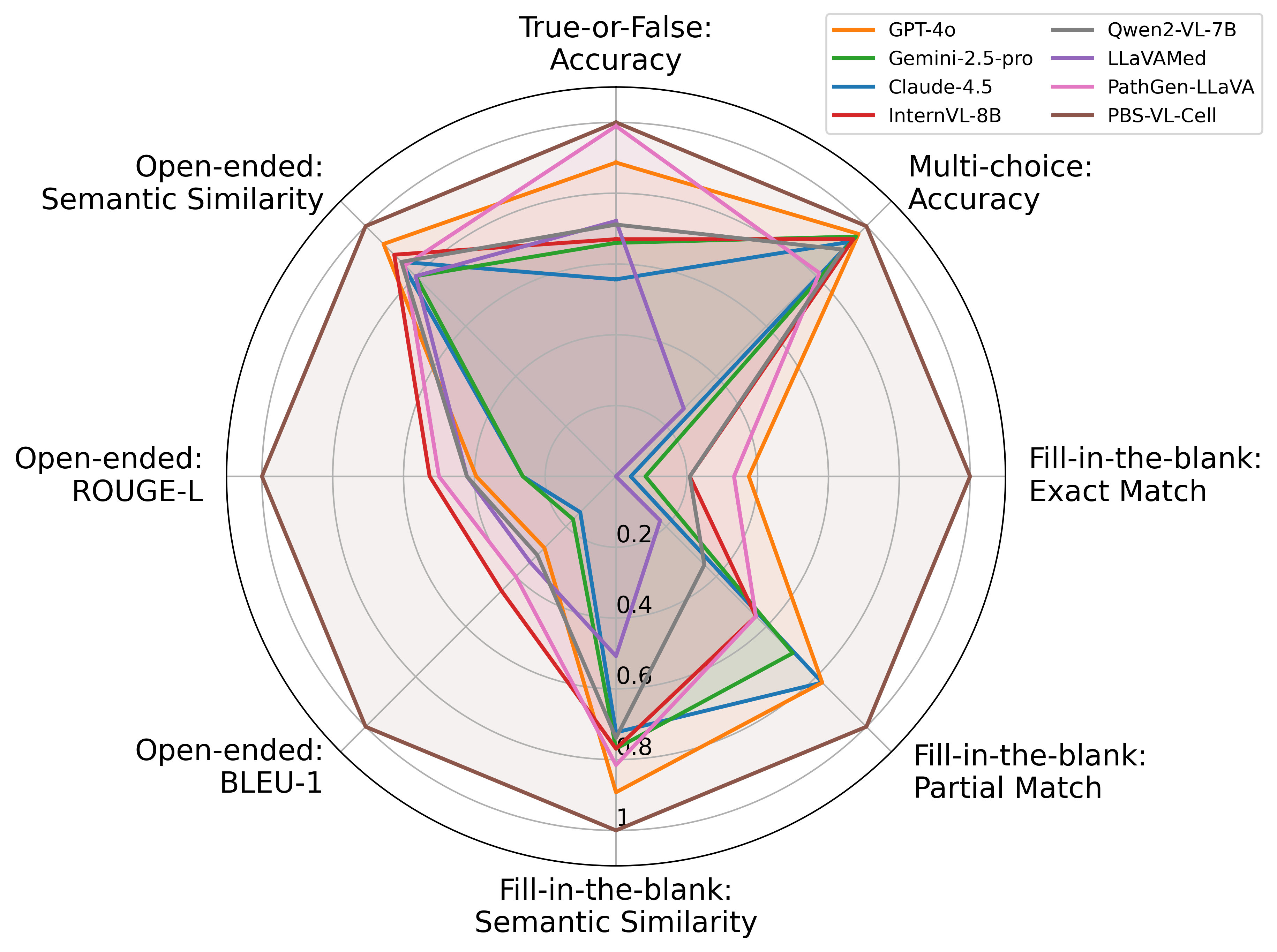}
    \caption{Out-of-domain evaluation for cell morphology QA}
    \label{fig:radar-a}
  \end{subfigure}
  \begin{subfigure}{0.46\linewidth}
    \includegraphics[width=\textwidth]{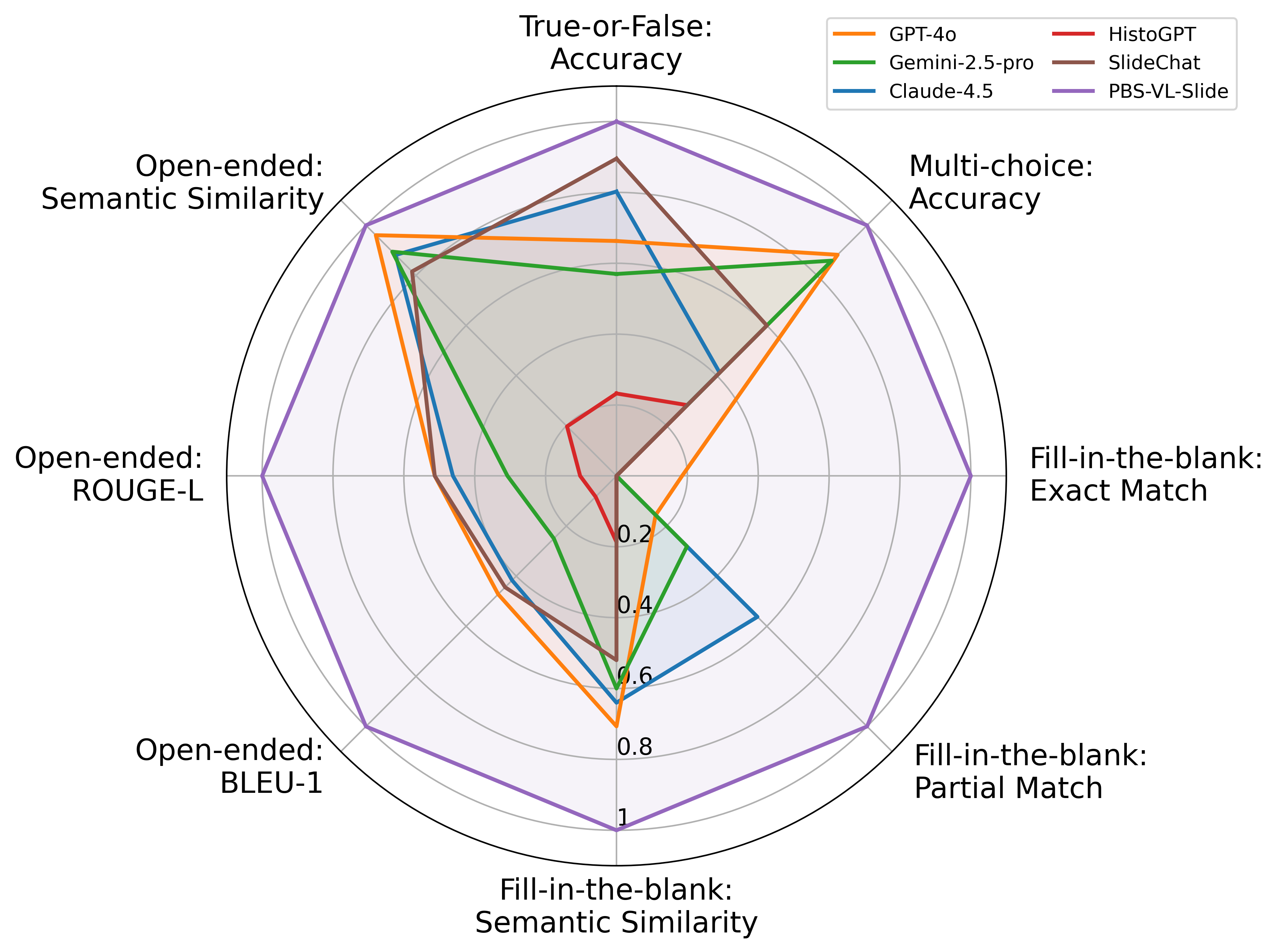}
    \caption{Evaluation for PBS whole slide understanding QA}
    \label{fig:radar-b}
  \end{subfigure}
  \caption{Illustration of the performance of existing models on PBSBench for both cell-level and slide-level QA. The metrics are normalized by the performance of the proposed model (PBS-VL).}
  \label{fig:radar}
\end{figure*}

Peripheral Blood Smear (PBS) is a fundamental hematopathology procedure and a key whole-slide imaging (WSI) modality for assessing the distributional and morphological abnormalities of blood cells. It plays a critical role in the screening of a broad spectrum of blood disorders, including leukemia, myelodysplastic syndrome, and anemia~\cite{pbs_intro}. As \cref{fig:wsi-comparison} shows, due to the fluid nature of blood, PBS WSIs have a significantly different appearance compared to solid tissue WSIs. In PBS slides, individual cells are distributed across a thin monolayer without tissue architecture, requiring pathologists to focus primarily on cell-level morphology rather than tissue organization. Consequently, while many multimodal large language models (MLLMs) are developed for computational pathology (CPath) on solid-tissue WSIs (\eg, TCGA)~\cite{Seyfioglu2023QuiltLLaVA, Chen2025SlideChat, Liang2025WSILLAVA}, these models are not well-suited for PBS interpretation.

Several challenges lie in developing MLLMs for PBS. First of all, publicly available data for PBS remains sparse compared to other tissues. While databases like The Cancer Genome Atlas Program (TCGA)~\cite{nci2025TCGA} contain over 10k WSIs for solid tissues, much fewer PBS WSIs are publicly accessible, making large-scale pretraining and evaluation difficult. Secondly, since WSIs are gigapixel-sized, the standard processing strategy is to tile them into patches. While the solid tissue occupies limited space in the WSI and produces empty patches that can be easily filtered out, PBS generates many more patches as blood spreads throughout the slide, resulting in an extreme computational cost and sparse, distributed abnormalities. Therefore, Rigorous quality control and patch selection are essential to build useful datasets. Finally, PBS interpretation centers on cytopathology, which means analyzing individual cells and their distribution, rather than the histopathology of tissue structures. This creates special needs for fine-grained, cell-level concentration in both model development and data curation.

In this paper, we tackle these challenges by developing a comprehensive multi-level vision-language framework for PBS interpretation. We first curate \textbf{PBSInstr}, a large-scale vision-language dataset comprising 353 PBS WSIs annotated with microscopic impression paragraphs. Additionally, we extract 29,673 cell-level image crops annotated with cell-type labels and detailed morphological descriptions, enabling fine-grained visual-textual alignment. We further curate 27,301 visual question-answer pairs on cell crops and 1,286 on PBS slides for instruction tuning. Upon PBSInstr, we tailor \textbf{PBS-VL}, a multi-scale vision-language model that integrates cell-level information into patch representation, then aggregates patch tokens into slide-level representations. To quantitatively evaluate the model's ability to interpret PBS, we further construct \textbf{PBSBench} using both in-domain data from PBSInstr and out-of-domain datasets. PBSBench is a dedicated visual question answering (VQA) benchmark encompassing four question types across six tasks, covering cell- and slide-level reasoning.

By evaluating existing MLLMs and PBS-VL on PBSBench. It turns out that PBS-VL outperforms existing general-purpose and pathology MLLMs across all question categories. To the best of our knowledge, this is the first framework enabling systematic vision-language development and evaluation on hematopathology.

To summarize, our contributions are:
\begin{itemize}
    \item We introduce \textbf{PBSInstr}, a large-scale vision-language dataset for PBS interpretation, comprising WSIs and cell image crops with paired textual captions, as well as QA pairs for instruction tuning.
    \item We propose \textbf{PBSBench}, a novel VQA benchmark featuring four question types on six tasks designed to evaluate both cell-level and WSI-level reasoning in PBS.
    \item We develop \textbf{PBS-VL}, a PBS-specific vision-language model that integrates fine-grained understanding of cell morphology with holistic slide-level interpretation.
    \item We release our code, datasets, and model weights as a foundation for future research on vision-language modeling in hematopathology.
\end{itemize}

%% file: sec_cr/2_related_works.tex
\section{Related Works}
\label{sec:related_work}

\begin{figure*}
    \centering
    \includegraphics[width=0.99\linewidth]{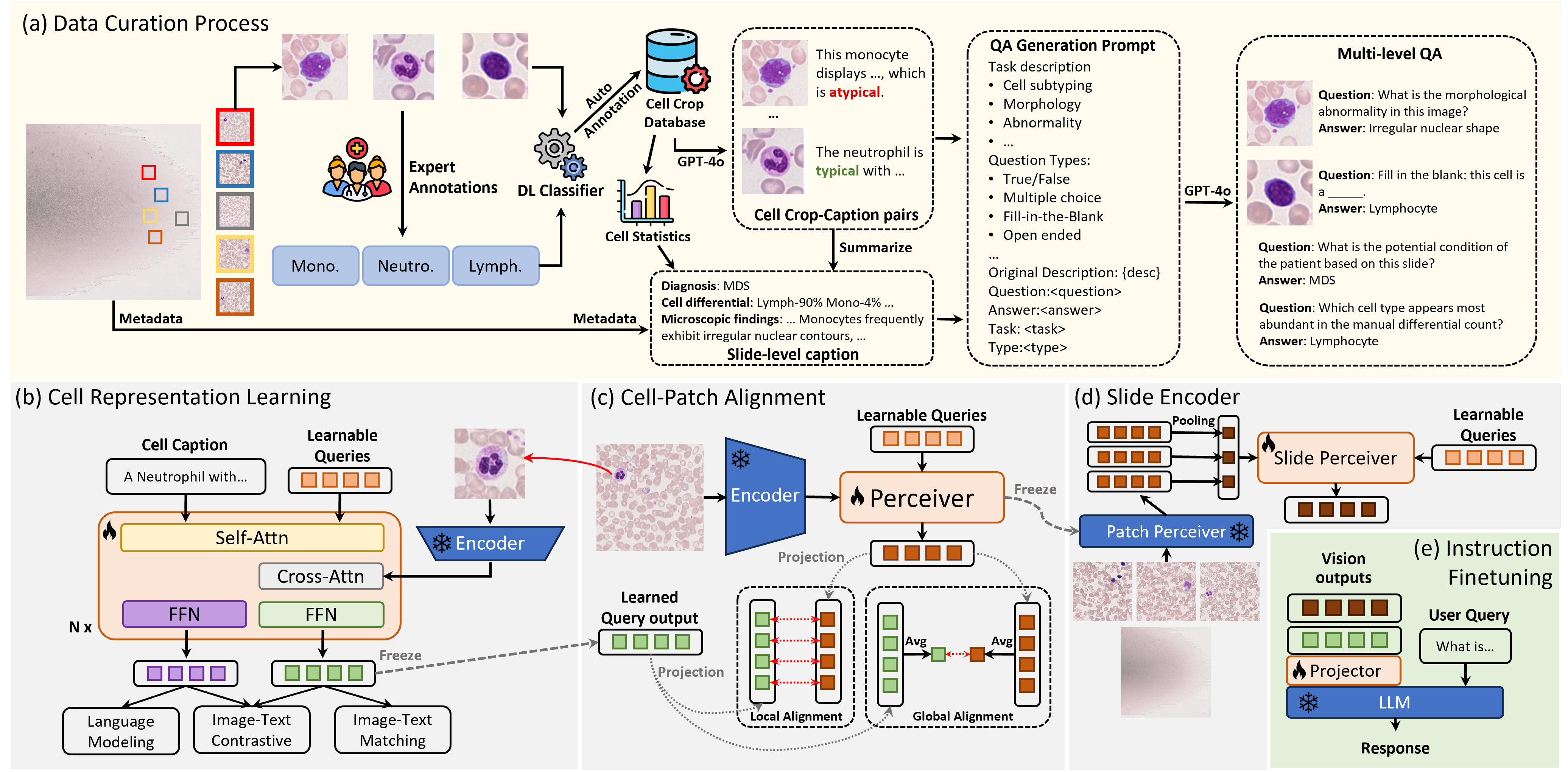}
    \caption{The Overview of our pipeline. }
    \label{fig:overview}
\end{figure*}

\subsection{Pathology QA Datasets}
\label{sec:related:datasets}

\paragraph{Textbooks and educational libraries.}
Early pathology VQA resources derive primarily from curated didactic materials. \textbf{PathVQA}~\cite{He2020PathVQA} compiles 32,799 open-ended Q\&A pairs over 4,998 images sourced from two widely used pathology textbooks and the PEIR digital library, offering the first broad benchmark for pathology-specific VQA beyond generic medical visual QA.

\paragraph{Educational video extraction.}
To scale image-text pairs, \textbf{Quilt-1M}~\cite{Ikezogwo2023Quilt1M} exploits publicly available YouTube histopathology teaching videos (1,087 hours) and additional web sources, yielding 1M image-text pairs for VL pretraining and retrieval. \textbf{PathMMU}~\cite{Sun2024PathMMU} further aggregates multiple sources: open-access articles, authoritative atlases/guidelines, expert social posts, public classification datasets, and YouTube educational content, and uses LMM-assisted curation and expert validation to produce 33,428 expert-level multiple choice questions over 24,067 images.

\paragraph{Whole-slide (WSI) datasets paired with reports.}
Recent WSI-centric benchmarks formulate slide-level reasoning from TCGA. TCGA-linked corpora of machine-readable reports~\cite{Kefeli2024TCGAReports} and WSI-report datasets (e.g., PathText~\cite{Chen2023WsiCaption}) further connect gigapixel slides with rich textual supervision. As for VQA, \textbf{WSI-VQA}~\cite{Chen2024WSIVQA} introduces a generative VQA framework and a curated WSI-QA corpus; in parallel, \textbf{SlideInstruction}/\textbf{SlideBench} (released with SlideChat~\cite{Chen2025SlideChat}) mines TCGA WSIs to provide 4.2K slide captions and 176K VQA pairs for instruction-following and evaluation. Most explicitly, \textbf{WSI-Bench} (released with WSI-LLaVA\citep{Liang2025WSILLAVA}) is curated directly from TCGA and finalizes 9,850 WSIs paired with their corresponding pathology reports across 30 cancer types, enabling morphology-aware WSI VQA at scale.

\subsection{Pathology Pretrained Models}
\label{sec:related:pretrained}

Given the gigapixel scale of WSIs, they are often tiled into small patches at a specific magnification level to fit modern vision models. Slide-level analysis is implemented by aggregating information from many local patches. As a result, pathology foundation models can be categorized into two main types: patch-level models and slide-level models.

\paragraph{Vision Foundation Models.}
Large self-supervised visual encoders trained on millions of histology tiles underpin many downstream tasks. \textbf{UNI}~\cite{Chen2024UNI} scales self-supervised training to 100K WSIs and 100M patches with strong transfer across 34 CPath tasks. \textbf{Virchow}~\cite{Vorontsov2024Virchow, Zimmermann2024Virchow2} trains ViT models using 1.5M WSIs, enabling robust pan-cancer detection and biomarker prediction. \textbf{GigaPath/Prov-GigaPath}~\cite{Xu2024ProvGigaPath} introduces long-context WSI modeling with open weights and a tile/slide dual-encoder pipeline. Additionally, \textbf{DinoBloom}~\cite{Koch2024DinoBloom} trains ViT models on a wide spectrum of blood cell images, laying a foundation for hematopathology analysis. From industry, \textbf{H-Optimus}~\cite{Bioptimus2024HOptimus0,Bioptimus2025HOptimus1} is developed on top of a proprietary collection of more than 500,000 H\&E-stained whole slide histology images.

\paragraph{Vision-language Foundation Models.}
Vision-language pretraining aligns pathology images with domain text. \textbf{PLIP}~\cite{Huang2023PLIP} demonstrates zero-shot classification and retrieval gains over generic CLIP. \textbf{CONCH}~\cite{Lu2024CONCH} scales their pretraining to 1.17M image-caption pairs and outperforms across 14 benchmarks (classification, segmentation, retrieval, captioning). \textbf{MUSK}~\cite{Xiang2025MUSK} employs unified masked modeling across tens of millions of images and billions of tokens, improving prognosis and precision-oncology tasks. These encoders are widely used as backbones or for zero- or few-shot transfer in pathology pipelines.

\paragraph{Slide-level foundation models.}
Slide-aware encoders often incorporate report text or slide-level supervision.  \textbf{PRISM}~\cite{Shaikovski2024PRISM} builds slide embeddings from Virchow tiles and clinical reports to support zero-shot cancer detection, report generation, and biomarker prediction. \textbf{CHIEF}~\cite{CHIEF2024} trains on 60,530 WSIs across 19 anatomic sites and generalizes to diverse external cohorts. \textbf{TITAN}~\cite{TITAN2025} combines slide-level SSL with vision-language alignment to 335,645 WSIs and synthetic captions. Collectively, these models bridge tile encoders and clinical slide-level reasoning.

\subsection{MLLMs for Pathology}

\paragraph{Patch-level MLLMs.} \textbf{Quilt-LLaVA}~\cite{Seyfioglu2023QuiltLLaVA} is trained with spatially localized instructions distilled from educational videos and improves open/closed-set VQA. \textbf{PathGen-LLaVA} relies on their constructed PathGen-1.6M dataset with a generated caption corpus to specialize LLaVA for pathology description and VQA~\cite{Sun2024PathGen16M}. \textbf{PA-LLaVA} provides a domain-specific assistant trained with pathology-aligned data and connectors~\cite{Dai2024PALLaVA}. \textbf{PathAsst}~\cite{Sun2024PathAsst} trains PathCLIP over 207K high-quality pathology image-text pairs, which serves as a dedicated encoder to enhance pathology analysis. \textbf{CPath-Omni}~\cite{Sun2025CpathOmni} integrates patch-level and slide-level analysis and solves tasks at different scales.

\paragraph{Whole-slide level MLLMs.}
Whole-slide assistants integrate slide encoders with LLMs. \textbf{WSI-VQA}~\cite{Chen2024WSIVQA} proposes generative VQA directly on WSIs. \textbf{SlideChat}~\cite{Chen2025SlideChat} provides an open-source assistant trained on TCGA WSI captions and 176K VQA pairs. \textbf{HistoGPT}~\cite{Tran2025HistoGPT} generates multi-section WSI reports validated by board-certified pathologists in dermatopathology. \textbf{WSI-LLaVA}~\cite{Liang2025WSILLAVA} highlights morphological understanding and explainability of model responses. The key to model development in WSI lies in effectively aggregating information from seas of patches.

%% file: sec_cr/3_method.tex
\section{Methodology}
\label{sec:method}

\begin{table*}[ht]
    \centering
    \resizebox{0.87\textwidth}{!}{%
\begin{tabular}{llllllll}
\toprule
      &       & \multicolumn{4}{c}{\# QA pairs by task}                          &              &              \\\cmidrule{3-6}
      & Split & Morphology & Abnormality & Cell subtyping    & Knowledge & Total \# QAs & \# cells     \\\midrule
\multirow{2}{*}{Cell level (ID)}    & Train & 13593      & 7533        & 3542              & 2633      & 27301        & 3744         \\
      & Test  & 1857       & 988         & 493               & 322       & 3660         & 500          \\\midrule
Cell level (OOD)   & -     & 2436       & 1258        & 147               & 654       & 4495         & 1498         \\\midrule
      &       & \multicolumn{5}{c}{\# QA pairs by task}                                         &              \\\cmidrule{3-7}
      & Split & Morphology & Abnormality & Cell differential & Knowledge & Diagnosis    & Total \# QAs \\\midrule
\multirow{2}{*}{Slide level (ID)} & Train & 642        & 47          & 305               & 138       & 154          & 1286         \\
      & Test  & 70         & 10          & 31                & 13        & 14           & 138         \\\bottomrule
\end{tabular}%
}
    \caption{The statistics of PBSInstr and PBSBench. PBSInstr contains QA from the training split at the cell and slide-level. PBSBench includes test split and OOD QAs. See \cref{sec:qa_gen} for details about QA tasks.}
    \label{tab:data_stat}
\end{table*}
The overview of our framework is shown in \cref{fig:overview}. The statistics of PBSInstr and PBSBench are shown in \cref{tab:data_stat}. More detailed statistics and all prompts used for data generation can be found in the supplementary materials.

\subsection{Data collection}
\label{sec:data_collection}

We started from S-BIAD440, a publicly available PBS slide dataset, curated for the Haemorasis study~\cite{De2023computational}. This dataset comprises 353 slides scanned at 40$\times$ magnification from patients with \textit{anemia} (69), \textit{myelodysplastic syndrome (MDS)} (227), and \textit{control} (57) conditions. We utilize these PBS WSIs to construct PBSInstr, a multi-level dataset comprising images with textual descriptions and question-answer pairs at both the single-cell and whole-slide levels.

\subsubsection{Patch tiling}
\label{sec:patch_tiling}

Medical whole-slide images are too large for modern vision models to process, and tiling is the most common method for handling them~\cite{TITAN2025, Tran2025HistoGPT, Chen2025SlideChat}. Therefore, we tile each WSI into non-overlapping 512$\times$512 patches at the native 40$\times$ magnification. Due to the fluid nature of blood films, raw tiling produces many low-quality tiles, including overcrowded cell fields, defocus blur, stain precipitate, and nearly empty tiles. To mitigate this, we adopt the tile quality control (QC) procedure from Haemorasis~\cite{De2023computational} to filter tiles. 

\subsubsection{Cell image extraction}
\label{sec:cell_extraction}

Hematopathology primarily focuses on cellular morphology, so we first extract cells from QC-approved tiles. We run Cellpose-SAM~\cite{Pachitariu2025CellposeSAM} on filtered tiles to produce instance masks. After that, we discard incomplete border instances and generate square crops centered on each instance, along with a small context of surrounding cells, to present the cell neighborhood. 

To obtain more information on cell crops, we categorize them by cell types with a two-stage strategy, which mitigates the extreme sample imbalance between red blood cells (RBCs) and white blood cells (WBCs). We begin by sampling and labeling 20,000 crops and train a lightweight binary classifier that identifies WBCs. Next, with the help of hematopathologists, we keep only WBCs and annotate them with fine-grained cell-type labels, including Basophils, Eosinophils, Lymphocytes, Monocytes, and Neutrophils. We also attach optional keyword comments that describe abnormal morphologies of the cell (\eg, hypolobated nuclei). To enhance comprehensiveness, we include crops containing artifacts, such as damaged cells, partially cropped cells, or staining artifacts. After removing unidentifiable crops, we got 1,567 human-labeled WBC crops. To further control error, multiple experienced pathologists reviewed the annotations to ensure consistency. Using this vetted set, we trained another classifier to annotate additional cell crops, yielding 28,106 more cell crops with confidence above the threshold. On a held-out subset of representative 300 crops, machine annotations achieved over 90\% accuracy under human review.

\subsection{PBSInstr}

\subsubsection{Morphological caption curation}
\label{sec:caption_curation}

Given the WBC crops with their respective cell types, we curate morphology-focused captions to expand the textual context. We prompt GPT-4o with three inputs: the cropped cell image, the annotated cell type, and optional keyword comments. It is instructed to identify observable abnormalities (\eg, over-/undersized cells, hyper-/hypolobated nuclei, membrane damage) and to suggest plausible underlying conditions in a non-committal manner when appropriate. For normal cells, a neutral description of salient visual appearance is produced. A stratified subset of 200 samples undergoes human review and 94\% of them are acceptable.

Beyond the single-cell observations, we also synthesize a structured caption for whole slides. It comprises (a) the recorded patient diagnosis from metadata, (b) slide-level WBC differentials estimated from our cell-crop classifier, and (c) a brief summary that conclude findings across the slide with emphasis on abnormalities. We obtain (c) by prompting GPT-4o to summarize the set of cell captions from the slide while preserving factual consistency. These curated captions then serve as the source for downstream question-answer (QA) synthesis.

\subsubsection{Question-answer generation}
\label{sec:qa_gen}

To unlock the hematopathology question-answering ability of MLLMs, we construct VQA pairs from image-caption pairs. We use four question types: \textbf{True-or-False} Judgement (binary fact checking), \textbf{Multiple-Choice Questions (MCQ)} (one correct option with distractors), \textbf{Fill-in-the-Blank} questions (keyword answers less than 10 words), and \textbf{Open-ended} questions (short sentence answers). 

Moreover, we target six tasks spanning multiple image scales and knowledge needs:

\begin{itemize}
    \item \textbf{Cell subtyping}: identify the cell type from a single-cell crop.
    \item \textbf{Diagnosis}: identify the underlying patient condition with whole-slide context.
    \item \textbf{Differential estimation}: estimate the slide-level WBC differential using classifier estimation as ground truth.
    \item \textbf{Morphology recognition}: detect morphological characteristics in the cell crops or the whole slide.
    \item \textbf{Abnormality detection}: name morphological abnormalities in the cell crops or whole slides. 
    \item \textbf{Knowledge mastery}: assess background knowledge relating morphology to clinical conditions.
\end{itemize}

Note that while knowledge mastery is not explicitly related to visual evidence, we include it as complement for learning essential domain knowledge in PBS understanding where the visual inputs can be seen as a problem context. They take around 10\% of questions.

To increase textual variety and reduce shortcuts, a given task can be presented in various formats (e.g., cell subtyping as MCQ, Fill-in-the-Blank, or Open-ended). We also randomize the order of MCQ options and constrain Open-ended answers to be brief, encouraging concise responses.

\begin{table*}[ht]
\centering
\caption{Cell Level Evaluation on PBSBench. Numbers in parentheses denote standard deviation by bootstrapping.}
\label{tab:cell-level-eval}
\resizebox{\textwidth}{!}{%
\begin{tabular}{l|cc|cc|ccc|cc|cc|ccc}
\toprule
                     & \multicolumn{7}{|c|}{In-domain}                                                                               & \multicolumn{7}{c}{Out   of Domain}                                                                         \\\midrule
                     & True/False  & Mul-Choice  & \multicolumn{2}{c|}{Fill in the Blank} & \multicolumn{3}{c|}{Open-ended QA}       & True/False  & Mul-Choice  & \multicolumn{2}{c|}{Fill in the Blank} & \multicolumn{3}{c}{Open-ended QA}       \\
                     & Accuracy         & Accuracy         & EMatch            & PMatch            & BLEU-1      & ROGUE-L     & Sim         & Accuracy         & Accuracy         & EMatch            & PMatch            & BLEU-1      & ROGUE-L     & Sim         \\\midrule
                     & \multicolumn{14}{c}{Proprietary VL Models}                                                                                                                                                                                \\\midrule
GPT-4o               & \textbf{0.77}(0.013)                    & 0.90(0.009)             & 0.23(0.014)   & 0.48(0.017)   & 0.09(0.002) & 0.14(0.002) & 0.65(0.005)  & 0.86(0.013)                    & 0.93(0.007)             & 0.09(0.008)   & 0.28(0.013)   & 0.10(0.001) & 0.15(0.002) & 0.65(0.004) \\
Gemini-2.5-pro       & 0.70(0.015)                    & 0.88(0.010)             & 0.03(0.006)   & 0.41(0.016)   & 0.06(0.001) & 0.10(0.001) & 0.60(0.005)  & 0.64(0.018)                    & 0.92(0.007)             & 0.02(0.004)   & 0.24(0.013)   & 0.06(0.001) & 0.10(0.001) & 0.56(0.004) \\
Claude-4.5-Sonnet    & 0.59(0.016)                    & 0.72(0.014)             & 0.02(0.005)   & \textbf{0.54}(0.016)   & 0.04(0.001) & 0.08(0.002) & 0.50(0.009)  & 0.54(0.018)                    & 0.90(0.008)             & 0.01(0.003)   & 0.28(0.014)   & 0.05(0.000) & 0.10(0.001) & 0.60(0.004) \\\midrule
                     & \multicolumn{14}{c}{Open-source General-purpose VL Model}                                                                                                                                                                 \\\midrule
InternVL-8B          & 0.73(0.014)                    & 0.91(0.009)             & 0.14(0.011)   & 0.38(0.016)   & 0.13(0.003) & 0.18(0.003) & 0.63(0.006)  & 0.65(0.017)                    & 0.91(0.007)             & 0.05(0.007)   & 0.19(0.012)   & 0.16(0.002) & 0.20(0.002) & 0.62(0.005) \\
Idefics3-8B          & 0.59(0.016)                    & 0.35(0.015)             & 0.11(0.010)   & 0.17(0.012)   & 0.14(0.003) & 0.17(0.004) & 0.56(0.007)  & 0.56(0.019)                    & 0.44(0.013)             & 0.03(0.005)   & 0.12(0.010)   & 0.14(0.003) & 0.18(0.003) & 0.56(0.005) \\
Qwen2-VL-7B          & 0.67(0.015)                    & 0.81(0.012)             & 0.16(0.012)   & 0.27(0.014)   & 0.10(0.003) & 0.14(0.003) & 0.62(0.005)  & 0.69(0.017)                    & 0.87(0.008)             & 0.05(0.007)   & 0.12(0.010)   & 0.11(0.002) & 0.16(0.002) & 0.60(0.005) \\
Qwen2.5-VL-7B        & 0.63(0.015)                    & 0.76(0.013)             & 0.03(0.006)   & 0.17(0.012)   & 0.04(0.001) & 0.07(0.001) & 0.59(0.005)  & 0.77(0.015)                    & 0.84(0.009)             & 0.01(0.003)   & 0.10(0.009)   & 0.04(0.000) & 0.07(0.001) & 0.58(0.004) \\
DeepSeek-VL-7B       & 0.63(0.015)                    & 0.51(0.015)             & 0.00(0.000)   & 0.15(0.011)   & 0.02(0.000) & 0.04(0.000) & 0.52(0.007)  & 0.94(0.009)                    & 0.59(0.013)             & 0.00(0.000)   & 0.07(0.008)   & 0.03(0.000) & 0.04(0.000) & 0.51(0.005) \\
LLaVA-1.6-Mistral-7B & 0.59(0.016)                    & 0.52(0.015)             & 0.06(0.008)   & 0.10(0.011)   & 0.05(0.001) & 0.08(0.001) & 0.56(0.005)  & 0.74(0.017)                    & 0.58(0.012)             & 0.02(0.005)   & 0.09(0.009)   & 0.05(0.001) & 0.08(0.001) & 0.55(0.004) \\
LLaVA-1.6-Vicuna-7B  & 0.58(0.016)                    & 0.46(0.016)             & 0.07(0.008)   & 0.10(0.010)   & 0.04(0.001) & 0.07(0.001) & 0.54(0.005)  & 0.70(0.017)                    & 0.55(0.013)             & 0.03(0.005)   & 0.08(0.008)   & 0.04(0.001) & 0.07(0.001) & 0.54(0.004) \\
BLIP2-OPT-6.7B       & 0.51(0.016)                    & 0.20(0.012)             & 0.01(0.003)   & 0.21(0.013)   & 0.09(0.003) & 0.11(0.003) & 0.41(0.007)  & 0.43(0.019)                    & 0.22(0.010)             & 0.01(0.002)   & 0.21(0.013)   & 0.07(0.002) & 0.09(0.002) & 0.36(0.006) \\\midrule
                     & \multicolumn{14}{c}{Domain-specific - Medical VL Model}                                                                                                                                                                   \\\midrule
LLaVA-Med            & 0.54(0.016)                    & 0.29(0.014)             & 0.00(0.000)   & 0.10(0.010)   & 0.11(0.002) & 0.15(0.002) & 0.56(0.006)  & 0.70(0.017)                    & 0.26(0.011)             & 0.00(0.000)   & 0.06(0.007)   & 0.12(0.002) & 0.16(0.002) & 0.56(0.005) \\
Med-flamingo         & 0.45(0.016)                    & 0.27(0.013)             & 0.00(0.000)   & 0.06(0.008)   & 0.07(0.002) & 0.09(0.002) & 0.43(0.007)  & 0.25(0.016)                    & 0.26(0.012)             & 0.00(0.000)   & 0.04(0.006)   & 0.07(0.001) & 0.08(0.001) & 0.38(0.005) \\\midrule
                     & \multicolumn{14}{c}{Domain-specific - Pathology VL Model}                                                                                                                                                                 \\\midrule
PA-LLaVA             & 0.39(0.015)                    & 0.24(0.013)             & 0.00(0.000)   & 0.00(0.000)   & 0.01(0.000) & 0.02(0.001) & -0.05(0.002) & 0.02(0.005)                    & 0.00(0.000)             & 0.00(0.000)   & 0.00(0.000)   & 0.00(0.000) & 0.00(0.000) & 0.01(0.001) \\
Quilt-LLaVA          & 0.61(0.015)                    & 0.36(0.015)             & 0.09(0.009)   & 0.16(0.012)   & 0.09(0.002) & 0.13(0.002) & 0.59(0.006)  & \textbf{0.97}(0.007)                    & 0.39(0.013)             & 0.04(0.007)   & 0.11(0.009)   & 0.09(0.001) & 0.13(0.002) & 0.58(0.005) \\
PathGen-LLaVA        & 0.65(0.015)                    & 0.78(0.012)             & 0.15(0.012)   & 0.31(0.015)   & 0.14(0.002) & 0.19(0.003) & 0.63(0.006)  & 0.96(0.007)                    & 0.78(0.011)             & 0.08(0.008)   & 0.19(0.012)   & 0.14(0.002) & 0.19(0.002) & 0.59(0.005) \\\midrule
Ours - PBS-VL                 & 0.73(0.015)                    & \textbf{0.96}(0.006)             & \textbf{0.47}(0.016)   & \textbf{0.54}(0.017)   & \textbf{0.36}(0.006) & \textbf{0.40}(0.006) & \textbf{0.72}(0.006)  & \textbf{0.97}(0.007)                    & \textbf{0.96}(0.005)             & \textbf{0.24}(0.013)   & \textbf{0.34}(0.014)   & \textbf{0.35}(0.004) & \textbf{0.38}(0.005) & \textbf{0.70}(0.005)\\\bottomrule 
\end{tabular}%
}
\end{table*}

\subsection{PBSBench}

To evaluate VQA performance on peripheral blood smears, we introduce PBSBench, a benchmark derived from and complementary to PBSInstr. PBSBench has two tracks: in-domain (ID) and out-of-domain (OOD) evaluation. The ID track is an independent hold-out split from PBSInstr at both the cell and slide levels. On the other hand, the OOD track applies our QA generation pipeline on external cell crops. We adopt three public datasets by normalizing their cell type labels to our cell categories:

\textbf{AML-Cytomorphology\_LMU} ~\cite{Matek2019AMLLMU} contains 18,365 expert-labeled single-cell images taken from peripheral blood smears of 100 patients diagnosed with Acute Myeloid Leukemia (AML) and 100 controls.

\textbf{APL-kaggle}~\cite{Shenderov2020APLKaggle} contains 14,189 single-cell images from peripheral blood smears of 106 Acute Promyelocytic Leukemia (APL) patients. 

\textbf{LISC}~\cite{Lisc2022WBCLisc, rezatofighi2011automatic} consists of PBS patches from the peripheral blood of 8 normal subjects from 100 microscope slides. Using the segmentation annotation provided by experts, we obtain 257 cell crops.

To evaluate model QA performance, we apply different metrics by question type. For closed-ended questions (True-or-False and MCQ), we use accuracy to measure the percentage of correctly answered questions. We advocate deterministic and controllable metrics for free text answers to mitigate evaluator bias and randomness. Therefore, for Fill-in-the-blank questions, we use the lexical matching strategy~\cite{wang2023evaluating} with two criteria. Exact Matching (EMatch) means that the answer is considered correct if and only if it is the same as the ground truth answers. To cater to MLLMs that tend to reply longer sentences, we also use Partial Matching (PMatch), which accepts answers that contain ground truth as a substring. As for open-ended questions, we utilize traditional NLP metrics for machine generation, including BLEU-1~\cite{papineni2002bleu} and ROUGE-L~\cite{lin2004rouge}. Additionally, we incorporate the cosine similarity of sentence representations from Sentence Transformers~\cite {reimers2019SentenceBert} as a semantic measurement. We perform case-folding and normalize whitespace and special characters before comparison.

\begin{table*}[ht]
\centering
\caption{Slide level Evaluation. Numbers in parentheses denote standard deviation by bootstrapping.}
\label{tab:slide-level-eval}
\resizebox{0.7\textwidth}{!}{%
\begin{tabular}{cl|ccc|cc|l}
\toprule
\multicolumn{1}{l}{}  &          & GPT-4o      & Gemini-2.5-pro & Claude-4.5  & HistoGPT    & SlideChat   & Ours - PBS-VL        \\\midrule
T/F                   & Accuracy & 0.57(0.082) & 0.49(0.080)    & 0.69(0.075) & 0.20(0.064) & 0.77(0.069) & \textbf{0.86}(0.056) \\\midrule
MC                    & Accuracy & 0.75(0.057) & 0.73(0.058)    & 0.35(0.063) & 0.24(0.056) & 0.51(0.065) & \textbf{0.85}(0.046) \\\midrule
\multirow{2}{*}{FB}   & EMatch   & 0.05(0.045) & 0.00(0.000)    & 0.00(0.000) & 0.00(0.000) & 0.00(0.000) & \textbf{0.27}(0.097) \\
                      & PMatch   & 0.05(0.043) & 0.09(0.063)    & 0.18(0.082) & 0.00(0.000) & 0.00(0.000) & \textbf{0.32}(0.096) \\\midrule
\multirow{3}{*}{Open} & BLEU-1   & 0.17(0.011) & 0.09(0.008)    & 0.15(0.014) & 0.03(0.003) & 0.16(0.018) & \textbf{0.36}(0.021) \\
                      & ROGUE-L  & 0.20(0.012) & 0.12(0.008)    & 0.18(0.015) & 0.04(0.004) & 0.20(0.014) & \textbf{0.39}(0.027) \\
                      & Sim      & 0.73(0.028) & 0.68(0.026)    & 0.67(0.027) & 0.15(0.014) & 0.62(0.032) & \textbf{0.76}(0.022)\\\bottomrule
\end{tabular}%
}
\end{table*}

\subsection{Multi-scale Model Architecture}

As shown in \cref{fig:overview}, we designed a bottom-up, multi-scale framework that mirrors the hierarchical nature of PBS analysis: information is first distilled from single-cell crops, then aligned by patch representations, and finally aggregated across the whole slide. This framework offers a multi-level representation that enables different tasks to be conducted at various levels.

\subsubsection{Cell-level framework}

Following BLIP-2~\cite{li2023blip2}, we concatenate a frozen vision backbone with a learnable Q-former to obtain text-informative visual tokens for cell crops. Concretely, the cell crop is encoded by the frozen backbone into a sequence of visual tokens. The Q-former then queries these features and produces a fixed number of vision-language tokens. The Q-former is pretrained following BLIP-2.

After the representation learning, the Q-former is attached to a linear projection followed by a downstream frozen VLM. We fine-tune the Q-former and the projection in an instruction-tuning manner with our constructed QA dataset for visual question answering, which also follows the vanilla BLIP-2 recipe.

\subsubsection{Cell-Patch Alignment}
\label{sec:pcalign}

To bypass explicit cell segmentation overhead during inference on unseen WSIs, we bridge the cell-level and slide-level analysis with a cell-aware patch encoder. For each patch, we obtain patch visual tokens with a frozen vision backbone. After that, we train a Perceiver~\cite{jaegle2021perceiver} module to compress patch visual tokens into $N$ latent tokens and align them with the cell Q-former outputs from cell crops within the patch, which also have $N$ tokens.

To align patch and cell visual tokens, we map them to the same space using a linear mapping. Let $V_p^{(i)},V_c^{(i)},i=1,2,\cdots N$ denote the patch and cell tokens, we introduce global and local alignment objectives as:

\begin{align*}
&\mathcal{L}_{global}=\left\lVert\frac{1}{N}\sum_{i=1}^{N}(V_p^{(i)}-V_c^{(i)})\right\rVert \\
&\mathcal{L}_{local}=-\frac{1}{N}\sum_{i=1}^{N}\log\left(\frac{\exp({V_p^{(i)}\cdot V_c^{(i)}})}{\sum_j \exp({V_p^{(i)}\cdot V_c^{(j)}})}\right)
\end{align*}

We conduct alignment between each cell with its corresponding patch and skip patches without cells. This alignment enables a purely patch-based inference path that approximates the cell tokens, bypassing segmentation while preserving cell-level semantics.

\subsubsection{Slide-level framework}

In line with the common practice of WSI processing~\cite{TITAN2025, Tran2025HistoGPT, Chen2025SlideChat}, we aggregate patch information across the whole slide to enable slide-level analysis. Patch tokens are first aggregated to provide a conclusive feature vector of the patch. As a slide can contain a massive number of patches, we apply a second Perceiver resampler to distill salient information from patches into a fixed number of slide tokens. These slide tokens serve as the visual inputs to the frozen VLM, and we conduct instruction-tuning with our slide-level QA while updating only the resampler and projector.

\subsection{Training scheme}

We train the multi-scale system in four phases to progressively align cell, patch, and slide inputs while keeping the vision backbone and downstream VLM frozen for stability and efficiency.

\begin{enumerate}
    \item \textbf{Vision-language Representation learning} freeze visual backbone and train Q-former on cell crops and captions with BLIP-2~\cite{li2023blip2} phase-I losses: image-text contrastive loss, image-text match loss, and language modeling loss. 
    \item \textbf{Cell-Patch Alignment} freeze cell Q-former from Phase 1 and train Perceiver on patches by global and local alignment. Patch tokens and cell tokens are mapped into a shared space and optimized with the global and local alignment objectives described in \cref{sec:pcalign}.
    \item \textbf{Cell Instruction Tuning} attach cell Q-former from Phase 1 to a projector and VLM, then conduct instruction tuning on PBSInstr cell QA pairs. Only the cell Q-former and the projector are trained.
    \item \textbf{Slide Instruction Tuning} use frozen patch representations from phase 2, and a Perceiver as visual encoder, attach to the projector and VLM, then conduct instruction tuning on PBSInstr slide QA pairs. Only the Perceiver and the projector are trained.
\end{enumerate}

%% file: sec_cr/4_experiments.tex
\begin{figure*}
    \centering
    \includegraphics[width=0.83\linewidth]{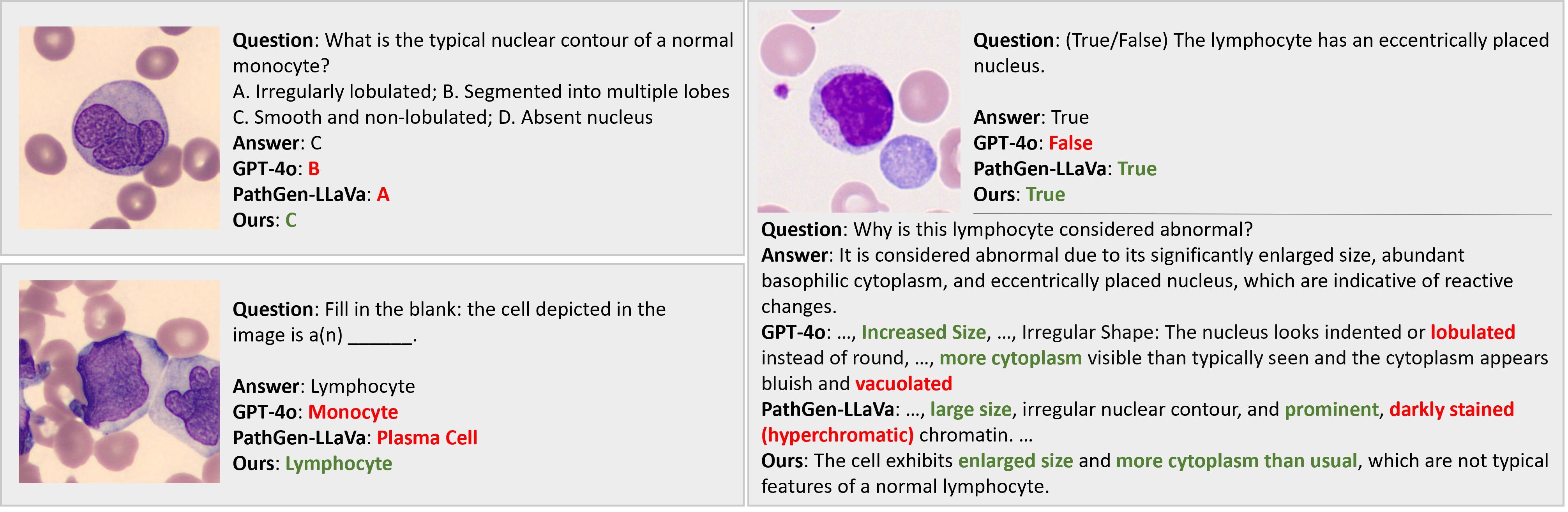}
    \caption{Case study of model performance of GPT-4o, PathGen-LLaVA, and our proposed model on each question type. We highlight the correct keywords in green and wrong keywords in red.}
    \label{fig:case_study}
\end{figure*}

\section{Experiments}
\label{sec:experiments}

\subsection{Baselines}

We include a wide spectrum of MLLMs across domains as baselines:

\begin{itemize}
    \item \textbf{Proprietary models}: GPT-4o~\cite{hurst2024gpt4o}, Claude Sonnet 4.5~\cite{anthropic2025Claude}, and Gemini 2.5 pro~\cite{google2025Gemini}
    \item \textbf{General-purpose models}: InternVL3.5 (8B)~\cite{wang2025internvl35}, Idefics3 (8B)~\cite{laurenccon2024idefics3}, Qwen2VL (7B)~\cite{wang2024qwen2}, Qwen2.5VL (7B)~\cite{bai2025qwen25}, DeepseekVLHybrid (7B)~\cite{lu2024deepseekvl}, LLaVA1.6 (7B)~\cite{liu2024llavanext}, and BLIP2-OPT (6.7B)~\cite{li2023blip2}
    \item \textbf{Medical-specific models}: LLaVA-Med~\cite{li2023LlavaMed} and Med-flamingo~\cite{moor2023MedFlamingo}
    \item \textbf{Pathology-specific models for patches}: PA-LLaVA~\cite{Dai2024PALLaVA}, Quilt-LLaVA~\cite{Seyfioglu2023QuiltLLaVA}, and PathGen-LLaVA
    \item \textbf{Pathology-specific models for WSIs}: HistoGPT~\cite{Tran2025HistoGPT} and SlideChat~\cite{Chen2025SlideChat}
\end{itemize}

\subsection{Implementation Details}

We choose Dinobloom ViT-L/14 as our vision backbone and Qwen 2.5 VL-7B as the backbone VLM. The number of query tokens is set to 32 for the cell Q-former, patch Perceiver, and slide Perceiver. We use a learning rate of 5e-5 with cosine scheduling and a 10\% warm-up period.

While benchmarking existing MLLMs, we disabled all sampling during the generation process and adhered to their chat template. For the benchmark on WSIs with proprietary models, we randomly sample 30 patches from each slide to provide the slide context~\cite{Chen2025SlideChat}. For WSI models, we follow their vision encoder and input format.

\subsection{Results}

\subsubsection{Cell-level Evaluation}

The result of the cell track benchmark is shown in \cref{tab:cell-level-eval}. It can be seen that the performance of existing models varied significantly in PBS understanding, and some models, such as BLIP-2, Med-Flamingo, and PA-LLaVA, performed near-randomly. While general-purpose models like GPT-4o, Gemini, and InternVL achieved remarkable results on closed-ended questions, their performance drops considerably for fill-in-the-blank and open-ended questions where no answer choices are available. This may reveal a potential drawback of these models: while they possess knowledge of hematopathology, they struggle to effectively associate textual knowledge with visual clues. As a result, they get the answer when given choices as hints, but fail when asked to come up with an answer directly. 

Additionally, domain-specific models, including those pathology-focused ones, do not substantially improve PBS understanding, highlighting the unique challenge of interpreting PBS. However, fine-tuning on PBSInstr yields significantly better performance than off-the-shelf models, and this improvement also generalizes to out-of-domain settings. Nevertheless, performance still drops on OOD data, especially for fill-in-the-blank questions that require precise, keyword-level answers.

\subsubsection{Slide-level Evaluation}


We report the model on slide-level performance in \cref{tab:slide-level-eval}. Generally, it shows a similar pattern to the cell-level. However, the models perform worse than cell crop interpretation, highlighting the greater complexity of slide analysis. Notably,  HistoGPT is primarily trained for report generation and struggles in chat-style QA. In contrast, our model still shows substantial gains after fine-tuning on PBSInstr, exhibiting its potential for improving PBS understanding.

\subsection{Case Study}

To further understand the performance difference, we conduct case studies of OOD cell crops. We chose GPT-4o for comparison, as it is our backbone LLM for creating the QA pairs. We also include PathGen-LLaVA as the best-performed domain-specific model. We present one sample from each question type with the model outputs in \cref{fig:case_study}. 

For the upper-right example, the monocyte has atypical nuclear contours, but the question asks about normal cells, which is a knowledge mastery task. GPT-4o and PathGen-LLaVA seem overly influenced by the image and answer incorrectly. In the lower-left example, the image contains multiple cells, but the target is the central one, which confuses PathGen-LLaVA. Also, GPT-4o fails to distinguish a monocyte from a reactive lymphocyte, which is a classic challenge in blood cell subtyping. The right-hand example similarly shows multiple objects that confuse the models. More observation from additional case studies shown in the supplementary material exhibits that GPT-4o performs better when the question contains the correct cell type. 

The case studies support our idea that the bottleneck of current MLLMs in understanding PBS may lie in establishing an accurate connection between visual clues and medical concepts. This observation further highlights the importance of PBSBench.

%% file: sec_cr/5_conclusion.tex
\section{Conclusion}
\label{sec:conclusion}

In this work, we present a multi-level vision-language framework and benchmark that focuses on understanding peripheral blood smears (PBS). We primarily focus on cytopathology and cytomorphology, which are currently underexplored in pathology studies. 

By introducing PBSInstr, PBSBench, and our PBS-VL model, we offer the first systematic evaluation of MLLMs on PBS at both the cell and slide levels, along with a multi-level model tailored to PBS interpretation. Our experiments show that although many existing MLLMs perform reasonably well on closed-ended questions, they often fail to correctly interpret PBS images from scratch, possibly because they cannot reliably extract key morphological concepts from visual evidence to leverage the rich medical knowledge already encoded in large language models. 

Looking ahead, we view this work as the first step toward comprehensive AI for hematopathology. Future models will need to handle a broader spectrum of WSIs across tissues and staining protocols, unifying cytopathology and histopathology. Through our work, we hope to highlight the importance of cytopathology-enriched approaches to complement existing models for WSI-focused systems and help bring trustworthy AI closer to hematopathology practice. 

%% file: sec_cr/A_data_stats.tex
\clearpage
\setcounter{page}{1}
\maketitlesupplementary

\section{Dataset Statistics}

Here in \cref{tab:supp_data_stat}, we present more detailed statistics on the number of questions in our datasets, with a breakdown by question type. Note that we remove questions with high similarity or trivial answers, leading to some small subgroups. 

\begin{table*}[ht]
\centering
\caption{Detailed breakdown of the number of questions w.r.t tasks and question types}
\label{tab:supp_data_stat}
\resizebox{0.8\textwidth}{!}{%
\begin{tabular}{ccl|lllll}
\toprule
\multicolumn{1}{l}{}           & \multicolumn{1}{l}{}      &                   & \multicolumn{5}{c}{\# QA pairs}                               \\\midrule
\multicolumn{1}{l}{Image type} & \multicolumn{1}{l}{Split} & Tasks             & True/False & MCQ  & Fill-in-the-Blank & Open-ended & Subtotal \\\midrule
\multirow{10}{*}{Cell level (ID)}    & \multirow{5}{*}{Train}    & Morphology        & 4972       & 4481 & 3263              & 877        & 13593    \\
                               &                           & Abnormality       & 1702       & 1729 & 808               & 3294       & 7533     \\
                               &                           & Cell subtyping    & 134        & 647  & 2360              & 401        & 3542     \\
                               &                           & Knowledge         & 334        & 1183 & 397               & 719        & 2633     \\
                               &                           & Subtotal          & 7142       & 8040 & 6828              & 5291       & 27301    \\\cmidrule{2-8}
                               & \multirow{5}{*}{Test}     & Morphology        & 670        & 625  & 454               & 108        & 1857     \\
                               &                           & Abnormality       & 231        & 222  & 103               & 432        & 988      \\
                               &                           & Cell subtyping    & 22         & 103  & 318               & 50         & 493      \\
                               &                           & Knowledge         & 34         & 138  & 41                & 109        & 322      \\
                               &                           & Subtotal          & 957        & 1088 & 916               & 699        & 3660     \\\midrule
\multirow{5}{*}{Cell level (OOD)}    & \multirow{5}{*}{-}        & Morphology        & 720        & 894  & 688               & 134        & 2436     \\
                               &                           & Abnormality       & 7          & 304  & 123               & 824        & 1258     \\
                               &                           & Cell subtyping    & 0          & 43   & 85                & 19         & 147      \\
                               &                           & Knowledge         & 0          & 252  & 179               & 223        & 654      \\
                               &                           & Subtotal          & 727        & 1493 & 1075              & 1200       & 4495     \\\midrule
\multirow{12}{*}{Slide level (ID)}   & \multirow{6}{*}{Train}    & Morphology        & 269        & 157  & 189               & 27         & 642      \\
                               &                           & Abnormality       & 16         & 21   & 4                 & 6          & 47       \\
                               &                           & Cell differential & 39         & 250  & 16                & 0          & 305      \\
                               &                           & Knowledge         & 0          & 1    & 0                 & 137        & 138      \\
                               &                           & Diagnosis         & 1          & 78   & 4                 & 71         & 154      \\
                               &                           & Subtotal \# QAs   & 325        & 507  & 213               & 241        & 1286     \\\cmidrule{2-8}
                               & \multirow{6}{*}{Test}     & Morphology        & 31         & 18   & 20                & 1          & 70       \\
                               &                           & Abnormality       & 0          & 6    & 1                 & 3          & 10       \\
                               &                           & Cell differential & 4          & 26   & 1                 & 0          & 31       \\
                               &                           & Knowledge         & 0          & 0    & 0                 & 13         & 13       \\
                               &                           & Diagnosis         & 0          & 5    & 0                 & 9          & 14       \\
                               &                           & Subtotal \# QAs   & 35         & 55   & 22                & 26         & 138     \\\bottomrule
\end{tabular}%
}
\end{table*}

\section{Code and Data Availability}

We publish our code for training \textbf{PBS-VL}\footnote{https://github.com/Wang-Yuanlong/PBSBench}. We also publish \textbf{PBSInstr}, \textbf{PBSBench}, and our evaluation toolkit together with the training code.

%% file: sec_cr/B_ethical_statement.tex
\section{Ethical, Limitation, and Hallucination Statements}

\subsection{Ethical Statement}

Our proposed datasets \textbf{PBSInstr} and \textbf{PBSBench} are primarily built on a publicly available PBS dataset~\cite{De2023computational} and several blood cell datasets~\cite{Matek2019AMLLMU, Shenderov2020APLKaggle, Lisc2022WBCLisc}. They are thoroughly de-identified and reveal no personal information. 

\textbf{PBSInstr}, \textbf{PBSBench}, and \textbf{PBS-VL} are released solely for research and educational purposes. They are not intended for, and must not be used as, a standalone diagnostic device or clinical decision-support system. Any clinical deployment of the would require additional prospective validation, rigorous safety and performance assessment across diverse patient populations.

\subsection{Limitation Statement}
\label{sec:limitation_stmt}

While our work presents resources for developing multimodal models for hematopathology, it comes with limitations that need to be considered:

\begin{itemize}
    \item Our work is built on public datasets for a specific patient cohort, with limited disease coverage and an unknown demographic distribution. It cannot reflect the variability in PBS appearance across institutions, scanners, staining protocols, or more hematologic conditions.
    \item Our data annotation is noisy despite human evaluation. The noise comes from the subjectivity of cell-type annotation and the potential hallucination from GPT-4o. We only conduct human cross-evaluation on small subsets, and it cannot entirely mitigate human subjectivity.
    \item We automatically estimate blood cell differentials from slides using cell subtyping models. This introduces additional model-induced noise. As a result, our reported performance on differential estimation tasks should be interpreted relative to this surrogate reference, not as absolute accuracy against expert quantification. 
\end{itemize}

\subsection{Close-loop Effect Discussion}

We acknowledge potential bias from GPT-assisted synthesis and mitigate it via (a) different context: synthesis uses extra context (\eg, cell type, condition) while evaluation is image-only, and (b) structured tasks (MCQ, FB) that are less style-sensitive. Empirically, GPT-4o does not trivially solve PBSBench and PBS-VL improves on structured subsets, suggesting gains beyond stylistic alignment.

\subsection{Hallucination Statement}

Our work involves two sources of potential hallucination: (1) hallucinations in GPT–assisted annotations used to construct \textbf{PBSInstr} and \textbf{PBSBench}, and (2) hallucinations from PBS-VL.

Our curation of \textbf{PBSInstr} and \textbf{PBSBench} relies on GPT-4o. It participated in the creation of cell crop captioning, slide description synthesizing, and QA pair creation. Although we tried our best to constrain the model responses by providing detailed instructions (as shown in \cref{sec:prompts}) and employed multiple human-in-the-loop strategies for error control, these responses may still be hallucinated and erroneous.

We develop \textbf{PBS-VL} via adapter fine-tuning based on Qwen2.5-VL~\cite{bai2025qwen25} with our vision backbone on our dataset. As a result, PBS-VL inherits potential hallucination from Qwen, our curated noisy dataset, and overfitting during fine-tuning. Potential hallucinations include statements that are not fully supported by the visual evidence or misinterpretation of hematologic findings.

%% file: sec_cr/C_data_processing.tex
\section{Additional Data Processing Details}
\subsection{Patch Quality Control}

\begin{figure*}
    \centering
    \includegraphics[width=0.85\linewidth]{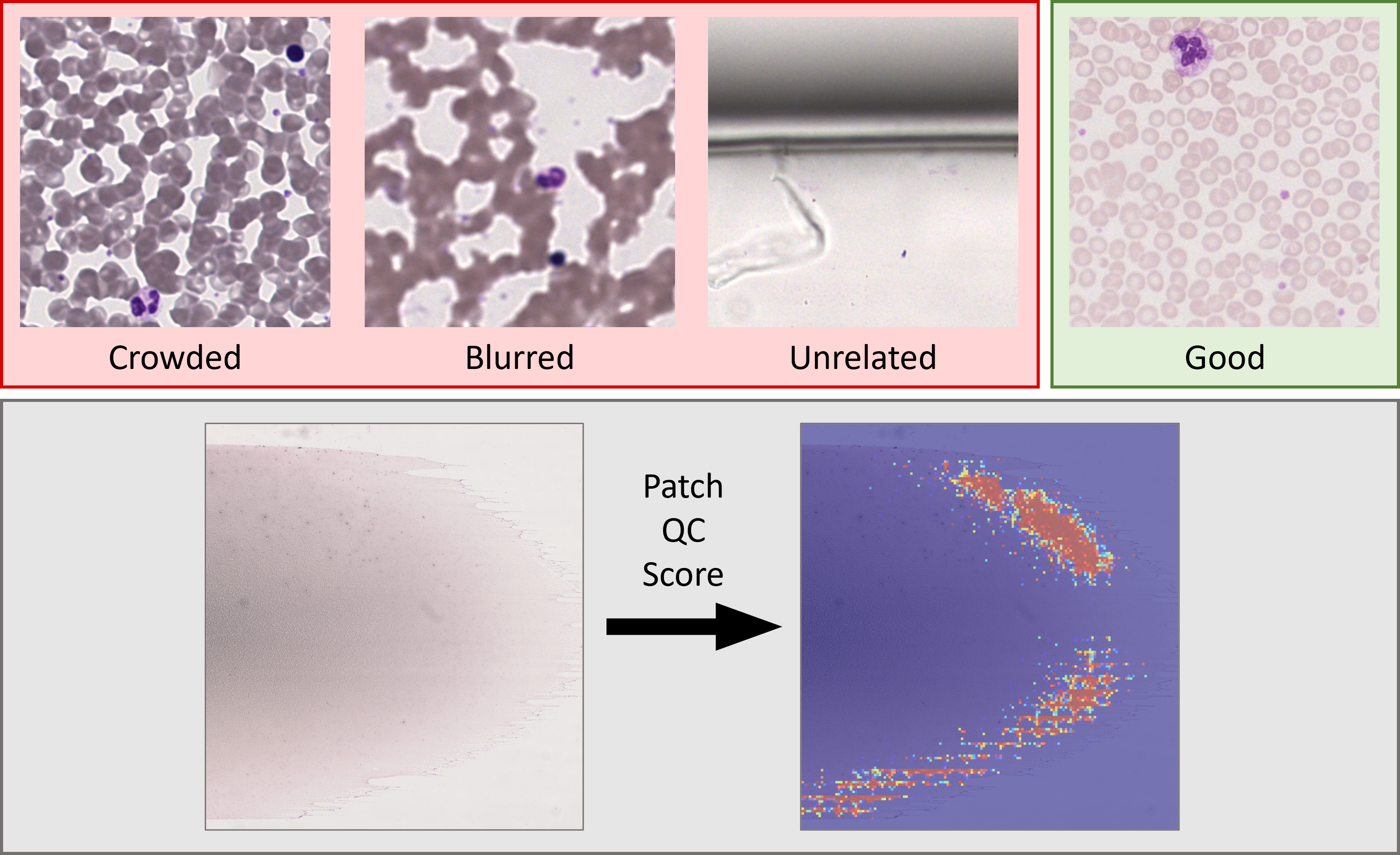}
    \caption{Quality control examples. Poor-quality patches are shown in the red box, and a good-quality patch is shown in the green box. Quality scores are shown in \textit{jet} colormap where red denotes high scores and blue denotes low scores.}
    \label{fig:qc}
\end{figure*}

We apply the tile quality control model from~\cite{De2023computational} to identify and remove patches with extremely low or high cell concentrations, as well as patches dominated by artifacts. Following~\cite{De2023computational}, we extract patches of size $512 \times 512$ at $40\times$ magnification and reuse their released DenseNet121-based quality control model~\cite{huang2017densenet} without additional fine-tuning. For each candidate patch, the model outputs a continuous quality score ranging from 0 to 1. We follow the original decision rule from~\cite{De2023computational} and set the quality score threshold to 0.5.

We illustrate typical quality control examples in \cref{fig:qc}. The upper panel shows examples of good- and poor-quality patches, highlighting standard failure modes such as overly crowded regions, edge patches without cells, and scanning artifacts. The lower panel exhibits the distribution of predicted quality scores across an entire slide, demonstrating how the model focuses on the monolayer region and the feathered edge area, which aligns with clinical practice.

\subsection{Cell Subtyping Model Details}

We develop a cell subtyping model based on Vision Transformer (ViT) variants from DinoBloom~\cite{Koch2024DinoBloom}. In particular, we use the ViT-L/14 (large) variant as the visual encoder. On top of the DinoBloom backbone, we append a feed-forward network (FFN) mapping head, followed by a linear classification layer. The FFN projects the global representation into a feature space tailored for downstream classification, and the final linear layer produces class logits. We freeze the DinoBloom backbone during training and update the FFN mapping head and the classifier. The models are trained with the Adam optimizer and a cosine learning rate schedule, using early stopping based on the area under the ROC curve (AUC) on a held-out validation set.

We train two separate models using expert-labeled PBS cell crops. The first is a binary classifier that distinguishes white blood cells (WBCs) from non-WBC components (\eg, red blood cells, platelets, and artifacts). The second is a multi-class WBC subtyping model that predicts fine-grained categories according to our predefined label set.

\subsection{Cell Type Mapping for OOD Datasets}

For out-of-distribution (OOD) evaluations, we normalize the cell type labels from external datasets to match our predefined categories. Specifically, we include three external datasets: AML-Cytomorphology\_LMU~\cite{Matek2019AMLLMU}, APL-Kaggle~\cite{Shenderov2020APLKaggle}, and LISC~\cite{Lisc2022WBCLisc}. The LISC dataset already uses the same WBC categories as our benchmark, so no relabeling is required. For AML-Cytomorphology\_LMU and APL-Kaggle, we map their original cell type to either a corresponding class in our label space or to a generic “Others” category when no match exists. The complete mapping between original labels and our standardized categories is shown in \cref{tab:cell_type_map}.

The “Others” category aggregates several types that would otherwise be too sparse or incompatible with our curated question–answer pairs. These include (1) rare WBC subtypes (\eg, myelocytes, giant thrombocytes), (2) cell types that do not appear in our cohort and thus cannot be consistently represented in our data (\eg, some blast categories), and (3) artifacts or unidentified labels that do not correspond to a well-defined cell category. This normalization step ensures consistent evaluation across datasets while avoiding unstable estimates for extremely rare or dataset-specific labels.

\subsection{VQA quality control}

We reject questions that are (a) non-PBS questions (\eg, general hematology), (b) factual errors, (c) not answerable from the image, and (d) answer-leaking (\ie, question implies the answer). These are also common error patterns observed from GPT-4o.

For labels mapped to Others, we exclude subtyping questions and ask morphology questions that remain visually grounded.

%% file: sec_cr/D_prompts.tex
\section{Prompts}
\label{sec:prompts}

As we scale up our annotation using GPT-4o, we present the prompt used for various tasks in \cref{fig:prompt_cell_caption}, \cref{fig:prompt_slide_caption}, \cref{fig:prompt_cell_qa}, and \cref{fig:prompt_slide_qa}. Note that we remove some prompt sections on input/output formatting and examples for simplicity.

%% file: sec_cr/E_additional_performance.tex
\section{Additional Experimental Results}

\subsection{Breakdown Performance on PBSBench}
In this section, we report a detailed breakdown of benchmark performance by task and question types. They can be found in \cref{tab:supp_result_tf}, \cref{tab:supp_result_mcq}, \cref{tab:supp_result_fb}, \cref{tab:supp_result_open}, and \cref{tab:supp_result_slide}. For cell-level questions, we report only one metric per question type for simplicity: accuracy for True-or-False and multiple-choice questions, partial-matching rate for Fill-in-the-blank, and semantic similarity for open-ended questions. Please note that some subgroups have a limited number of questions. As a result, the model's performance is subject to data randomness, making it unreliable to compare and interpret model performance on those subgroups.

\begin{figure*}[h]
    \centering
    \includegraphics[width=\linewidth]{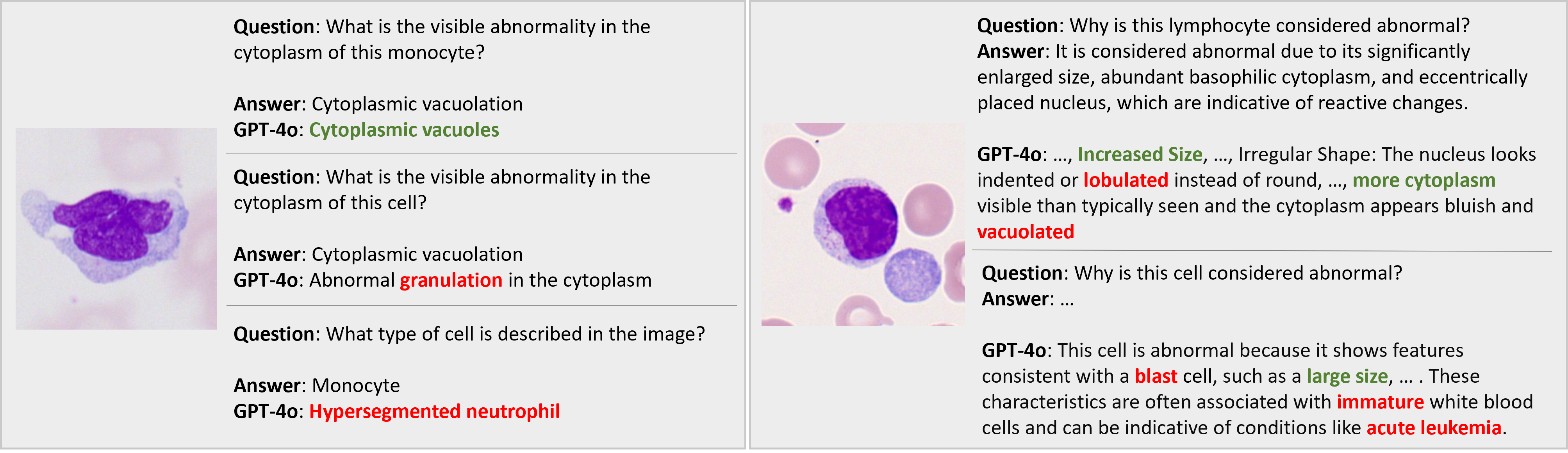}
    \caption{Additional case study on GPT-4o, the model shows performance degradation when no cell types are available as hints in questions.}
    \label{fig:supp_case_study}
\end{figure*}

\subsection{Sensitivity to the Number of Patches}

\begin{table}[ht]
\centering
\caption{Results of sensitivity of numbers of input patches for proprietary models Std in parenthesis}
\label{tab:supp_npatches}
\vspace{-0.15cm}
\resizebox{0.47\textwidth}{!}{%
\begin{tabular}{lccccc}
\toprule
model                              & \# patches & TF-Acc      & MCQ-Acc     & FB-Pmatch   & Open-BLEU1  \\\midrule
\multirow{3}{*}{GPT-4o}            & 15         & 0.63(0.077) & 0.80(0.048) & 0.09(0.060) & 0.18(0.015) \\
                                   & 30         & 0.57(0.077) & 0.75(0.057) & 0.05(0.045) & 0.17(0.011) \\
                                   & 50         & 0.71(0.071) & 0.84(0.049) & 0.18(0.086) & 0.18(0.013) \\\midrule
\multirow{3}{*}{Gemini-2.5-pro}    & 15         & 0.51(0.081) & 0.67(0.061) & 0.09(0.062) & 0.10(0.008) \\
                                   & 30         & 0.49(0.081) & 0.73(0.057) & 0.09(0.060) & 0.09(0.008) \\
                                   & 50         & 0.57(0.079) & 0.76(0.056) & 0.09(0.059) & 0.10(0.008) \\\midrule
\multirow{3}{*}{Claude-4.5-Sonnet} & 15         & 0.66(0.076) & 0.36(0.064) & 0.14(0.071) & 0.14(0.013) \\
                                   & 30         & 0.69(0.075) & 0.35(0.064) & 0.18(0.081) & 0.15(0.014) \\
                                   & 50         & 0.71(0.077) & 0.35(0.064) & 0.14(0.072) & 0.15(0.013)\\\bottomrule
\end{tabular}%
}
\end{table}

We conduct sensitivity analysis for the number of patches in \cref{tab:supp_npatches}. Note that GPT-4o allows up to 50 images. Results show that more patches can improve performance, but gains are modest and remain far from PBS-VL.

\subsection{Cell-patch Alignment Training Ablation}

\begin{table}[ht]
\centering
\caption{Results of cell-patch alignment training ablation on slide level QA. Std in parenthesis}
\label{tab:supp_ablation}
\vspace{-0.15cm}
\resizebox{0.47\textwidth}{!}{%
\begin{tabular}{lcccc}
\toprule
model                              & TF-Acc      & MCQ-Acc     & FB-Pmatch   & Open-BLEU1 \\\midrule
PBS-VL w/o align                                                    & \textbf{0.91}(0.047)                & 0.82(0.051)                 & 0.23(0.088)                   & 0.35(0.017)                    \\\midrule
PBS-VL                                                  & 0.86(0.056)                        & \textbf{0.85}(0.046)                 & \textbf{0.32}(0.096)                   & \textbf{0.36}(0.021)                   \\\bottomrule
\end{tabular}%
}
\end{table}

We conduct ablation of cell-patch alignment training by a variation with cell images as input in replacement of patch images for slide-level tasks. The result comparison is in \cref{tab:supp_ablation}. The original PBS-VL outperforms on MCQ, FB, and open-ended questions. Moreover, it bypasses cell detection at inference, reducing computational overhead.

%% file: sec_cr/F_case_study.tex
\section{Additional Case Study}

We provide an additional case study in \cref{fig:supp_case_study} that illustrates how model performance degrades when explicit cell type information is removed from the question. In this case study, we modify the original questions by replacing specific cell types (\eg monocyte) with a generic reference (“this cell”). Thus, we eliminate hints that may reveal or narrow down the answer without visual understanding. 

As shown in \cref{fig:supp_case_study}, GPT-4o can produce correct answers when the ground-truth cell type is stated in the question. However, it still struggles to correctly identify the cell type in the image on its own once textual hints are removed, leading to erroneous responses. This behavior supports our hypothesis that modern MLLMs possess substantial knowledge of hematopathology but still struggle to connect medical concepts to visual evidence.

%% file: sec_cr/G_large_tabels.tex
\newpage
\section{Large Tables and Figures}

See below.

\begin{table*}[!ht]
\centering
\caption{The mapping of cell type from out-of-distribution cell image datasets for normalization.}
\label{tab:cell_type_map}
\resizebox{0.77\textwidth}{!}{%
\begin{tabular}{ll|ll}
\toprule
\multicolumn{2}{c|}{AML-Cytomorphology\_LMU} & \multicolumn{2}{c}{APL-kaggle}             \\
fine-coarsed types    & normalized types    & fine-coarsed types      & normalized types \\\midrule
BAS                   & Basophil            & Artifact                & Others           \\
EBO                   & Others              & Band neutrophils        & Neutrophil       \\
EOS                   & Eosinophil          & Basophil                & Basophil         \\
KSC                   & Others              & Blast (no lineage spec) & Others           \\
LYA                   & Lymphocyte          & Eosinophils             & Eosinophil       \\
LYT                   & Lymphocyte          & Erythroblast            & Others           \\
MMZ                   & Others              & Giant thrombocyte       & Others           \\
MOB                   & Others              & Lymphocyte              & Lymphocyte       \\
MON                   & Monocyte            & Lymphocyte (variant)    & Lymphocyte       \\
MYB                   & Others              & Metamyelocyte           & Others           \\
MYO                   & Others              & Monocyte                & Monocyte         \\
NGB                   & Neutrophil          & Myelocyte               & Others           \\
NGS                   & Neutrophil          & Plasma cells            & Others           \\
PMB                   & Others              & Prolymphocyte           & Others           \\
PMO                   & Others              & Promonocyte             & Others           \\
                      &                     & Promyelocyte            & Others           \\
                      &                     & Segmented neutrophils   & Neutrophil       \\
                      &                     & Smudge cells            & Others           \\
                      &                     & Thrombocyte aggregation & Others           \\
                      &                     & Unidentified            & Others           \\
                      &                     & Young Unidentified      & Others       \\\bottomrule   
\end{tabular}%
}
\end{table*}

\begin{figure}[!ht]
    \centering
    \includegraphics[width=0.46\textwidth]{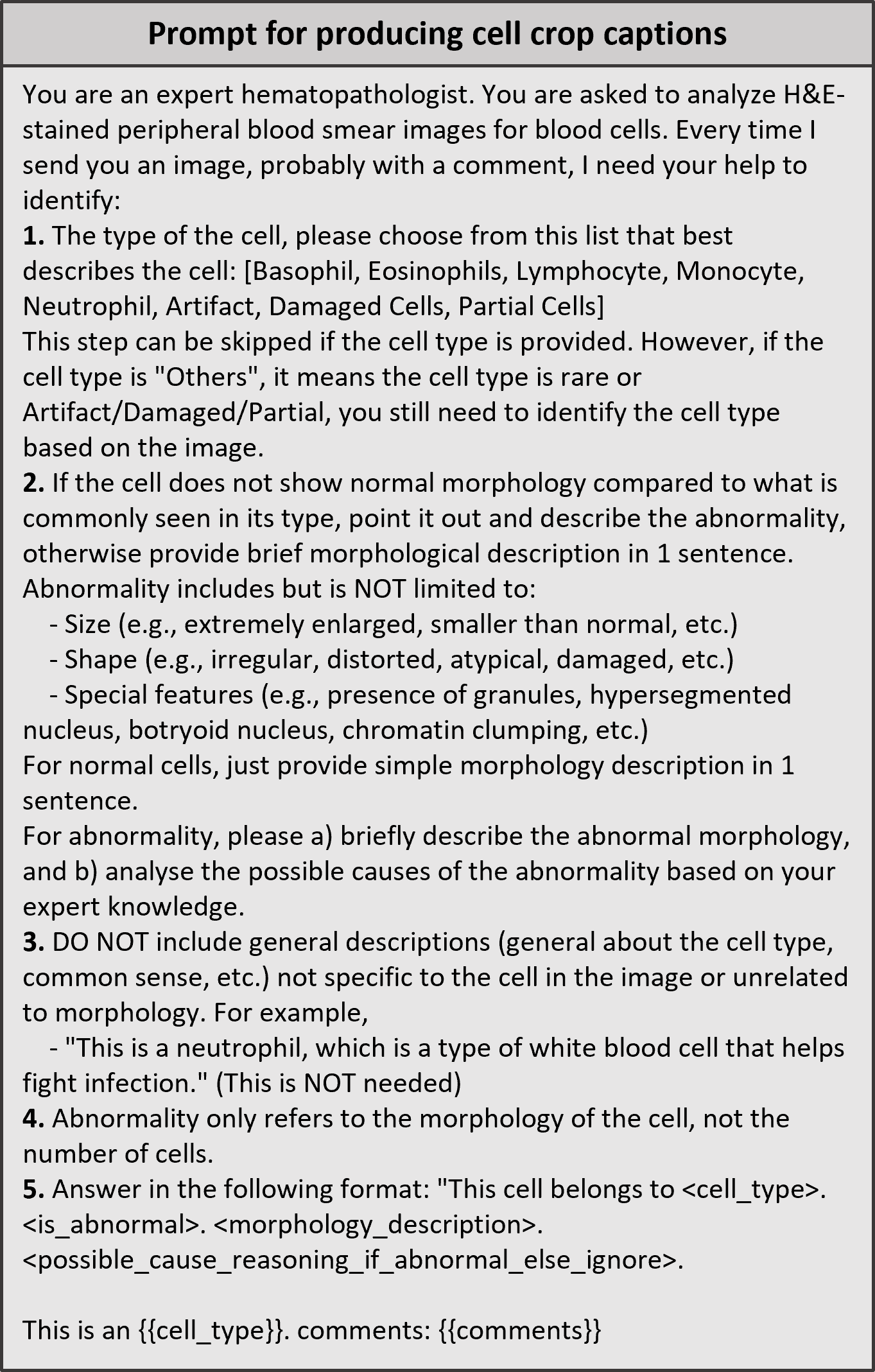}
    \caption{Prompts for cell crop captioning}
    \label{fig:prompt_cell_caption}
\end{figure}

\begin{figure}[!ht]
    \centering
    \includegraphics[width=0.46\textwidth]{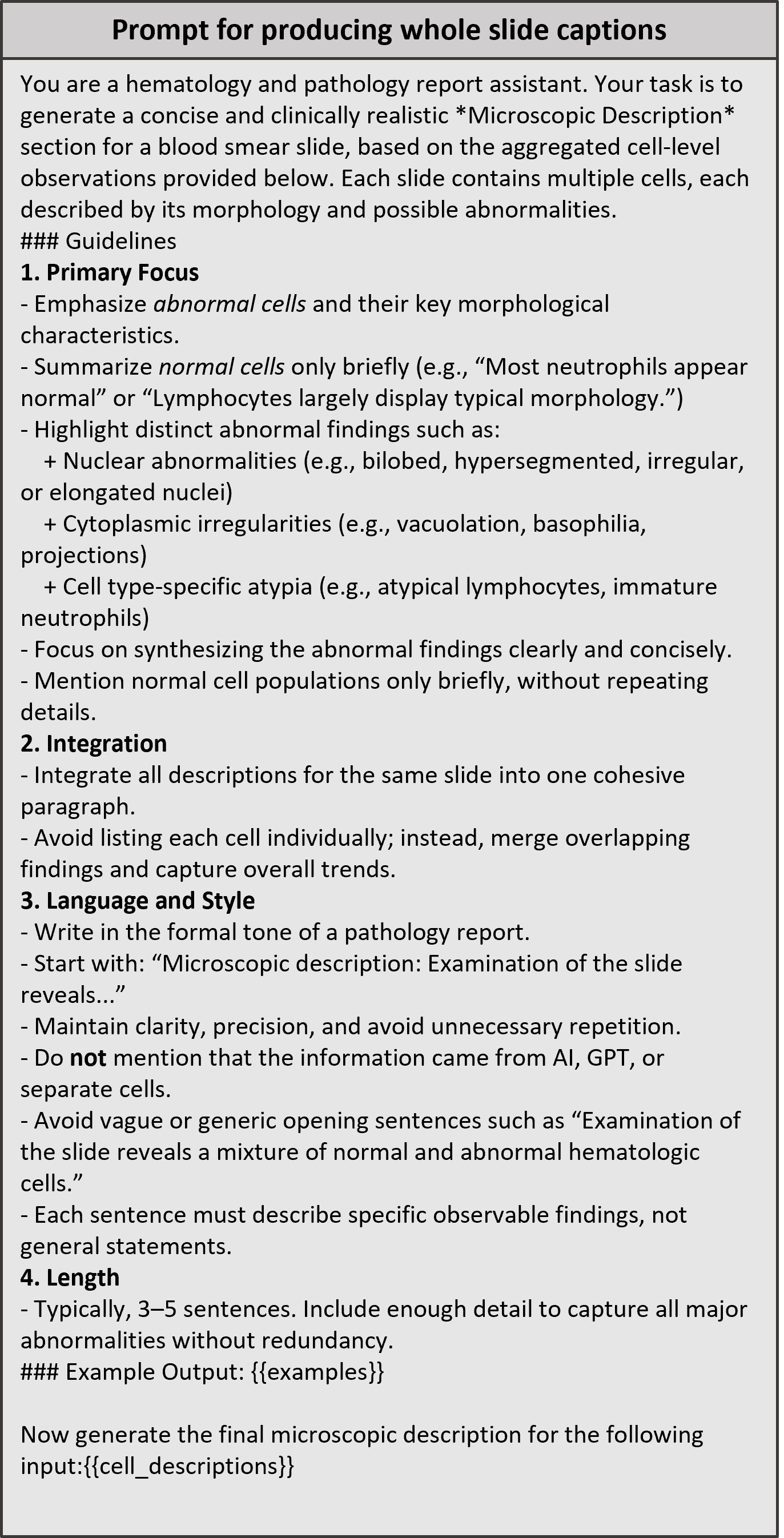}
    \caption{Prompts for whole slide captioning}
    \label{fig:prompt_slide_caption}
\end{figure}

\begin{figure}[!ht]
    \centering
    \includegraphics[width=0.46\textwidth]{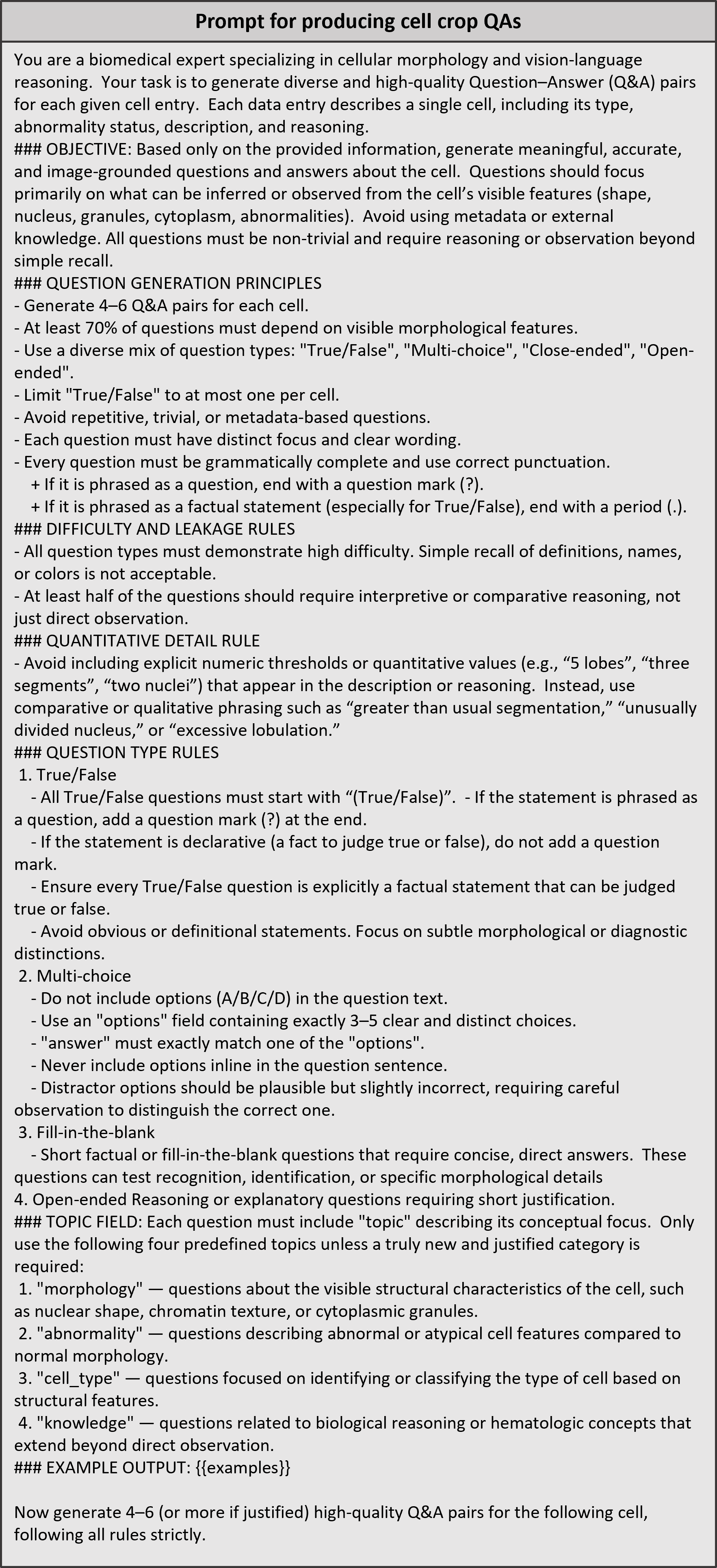}
    \caption{Prompts for cell crop QA}
    \label{fig:prompt_cell_qa}
\end{figure}

\begin{figure}[!ht]
    \centering
    \includegraphics[width=0.46\textwidth]{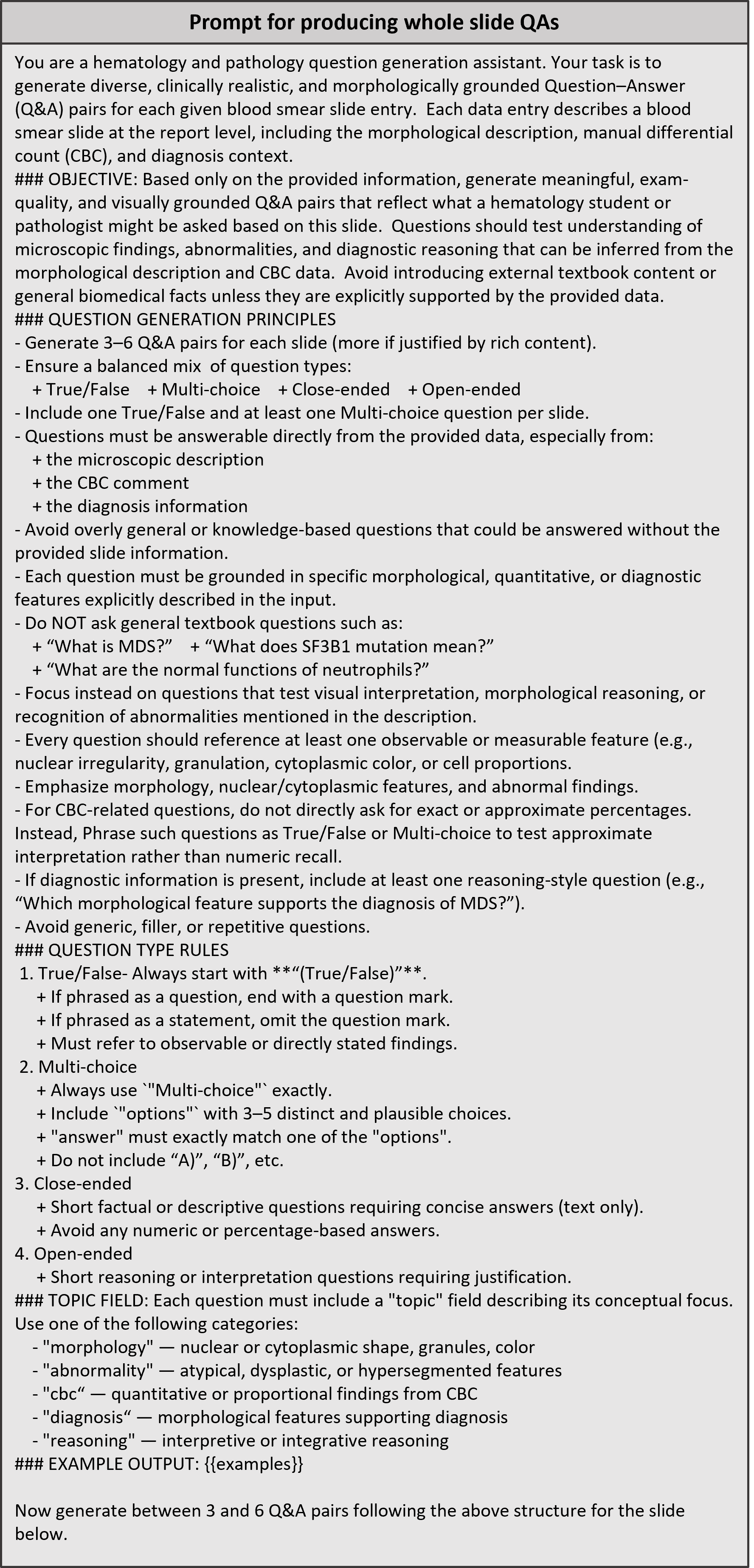}
    \caption{Prompts for whole slide QA}
    \label{fig:prompt_slide_qa}
\end{figure}

\begin{table*}[ht]
\centering
\caption{The accuracy of cell-level \textbf{True/False} questions for benchmarked models breaks down by task types.}
\label{tab:supp_result_tf}
\resizebox{0.85\textwidth}{!}{%
\begin{tabular}{l|cccc|cc}
\toprule
                     & \multicolumn{4}{|c|}{In-domain}                            & \multicolumn{2}{c}{Out of Domain} \\\cmidrule{2-7}
                     & Morphology  & Abnormality & Cell subtyping & Knowledge   & Morphology      & Abnormality     \\\midrule
                     & \multicolumn{6}{c}{Proprietary VL Models}                                                    \\\cmidrule{2-7}
GPT-4o               & 0.85(0.013) & 0.50(0.032) & 0.77(0.090)    & 0.91(0.048) & 0.86(0.013)     & 0.86(0.129)     \\
Gemini-2.5-pro       & 0.77(0.017) & 0.46(0.032) & 0.82(0.081)    & 0.94(0.040) & 0.65(0.018)     & 0.29(0.170)     \\
Claude-4.5-Sonnet    & 0.57(0.019) & 0.59(0.032) & 0.77(0.089)    & 0.89(0.053) & 0.54(0.019)     & 0.29(0.178)     \\\midrule
                     & \multicolumn{6}{c}{Open-source General-purpose VL Model}                                     \\\cmidrule{2-7}
InternVL-8B          & 0.75(0.016) & 0.65(0.031) & 0.91(0.059)    & 0.80(0.066) & 0.64(0.018)     & 0.86(0.136)     \\
Idefics3-8B          & 0.57(0.019) & 0.62(0.032) & 0.59(0.106)    & 0.77(0.069) & 0.56(0.019)     & 0.29(0.178)     \\
Qwen2-VL-7B          & 0.71(0.017) & 0.53(0.034) & 0.68(0.098)    & 0.80(0.068) & 0.69(0.016)     & 0.57(0.190)     \\
Qwen2.5-VL-7B        & 0.71(0.017) & 0.40(0.031) & 0.68(0.097)    & 0.80(0.069) & 0.78(0.015)     & 0.57(0.184)     \\
DeepSeek-VL-7B       & 0.77(0.016) & 0.24(0.028) & 0.45(0.105)    & 0.66(0.081) & 0.94(0.008)     & 1.00(0.000)     \\
LLaVA-1.6-Mistral-7B & 0.66(0.018) & 0.38(0.033) & 0.45(0.104)    & 0.80(0.067) & 0.75(0.016)     & 0.29(0.172)     \\
LLaVA-1.6-Vicuna-7B  & 0.64(0.018) & 0.39(0.032) & 0.68(0.101)    & 0.57(0.084) & 0.70(0.017)     & 0.43(0.185)     \\
BLIP2-OPT-6.7B       & 0.49(0.019) & 0.57(0.032) & 0.50(0.112)    & 0.54(0.083) & 0.43(0.018)     & 0.29(0.166)     \\\midrule
                     & \multicolumn{6}{c}{Domain-specific - Medical VL Model}                                       \\\cmidrule{2-7}
LLaVA-Med            & 0.60(0.018) & 0.35(0.031) & 0.55(0.105)    & 0.66(0.082) & 0.70(0.017)     & 0.71(0.170)     \\
Med-flamingo         & 0.38(0.019) & 0.65(0.031) & 0.32(0.096)    & 0.46(0.085) & 0.25(0.016)     & 0.43(0.185)     \\\midrule
                     & \multicolumn{6}{c}{Domain-specific - Pathology VL Model}                                     \\\cmidrule{2-7}
PA-LLaVA             & 0.26(0.017) & 0.79(0.027) & 0.50(0.103)    & 0.29(0.077) & 0.02(0.005)     & 0.14(0.133)     \\
Quilt-LLaVA          & 0.73(0.018) & 0.23(0.029) & 0.55(0.105)    & 0.71(0.074) & 0.97(0.006)     & 0.86(0.137)     \\
PathGen-LLaVA        & 0.78(0.016) & 0.25(0.027) & 0.73(0.094)    & 0.80(0.068) & 0.97(0.007)     & 0.71(0.170)     \\
Ours - PBS-VL        & 0.87(0.013) & 0.30(0.029) & 1.00(0.000)    & 0.83(0.063) & 0.97(0.007)     & 0.86(0.124)    \\\bottomrule
\end{tabular}%
}
\end{table*}

\begin{table*}[ht]
\centering
\caption{The accuracy of cell-level \textbf{MCQ} questions for benchmarked models breaks down by task types.}
\label{tab:supp_result_mcq}
\resizebox{0.9\textwidth}{!}{%
\begin{tabular}{l|cccc|cccc}
\toprule
                     & \multicolumn{4}{|c|}{In-domain}                            & \multicolumn{4}{c}{Out of Domain}                        \\\cmidrule{2-9}
                     & Morphology  & Abnormality & Cell subtyping & Knowledge   & Morphology  & Abnormality & Cell subtyping & Knowledge   \\\midrule
                     & \multicolumn{8}{c}{Proprietary VL Models}                                                                           \\\cmidrule{2-9}
GPT-4o               & 0.88(0.012) & 0.87(0.022) & 0.94(0.023)    & 0.99(0.007) & 0.91(0.010) & 0.95(0.013) & 0.95(0.031)    & 0.96(0.011) \\
Gemini-2.5-pro       & 0.85(0.015) & 0.87(0.023) & 0.94(0.023)    & 0.99(0.007) & 0.89(0.010) & 0.93(0.015) & 0.95(0.032)    & 0.99(0.007) \\
Claude-4.5-Sonnet    & 0.73(0.018) & 0.66(0.033) & 0.79(0.042)    & 0.74(0.038) & 0.87(0.011) & 0.91(0.016) & 0.98(0.023)    & 0.97(0.011) \\\midrule
                     & \multicolumn{8}{c}{Open-source General-purpose VL Model}                                                            \\\cmidrule{2-9}
InternVL-8B          & 0.89(0.013) & 0.93(0.015) & 0.93(0.025)    & 0.98(0.012) & 0.88(0.011) & 0.94(0.014) & 0.95(0.032)    & 0.97(0.010) \\
Idefics3-8B          & 0.30(0.018) & 0.34(0.032) & 0.32(0.047)    & 0.62(0.041) & 0.34(0.016) & 0.54(0.028) & 0.26(0.068)    & 0.68(0.029) \\
Qwen2-VL-7B          & 0.74(0.017) & 0.92(0.018) & 0.81(0.040)    & 0.97(0.014) & 0.82(0.012) & 0.93(0.015) & 0.84(0.056)    & 0.97(0.011) \\
Qwen2.5-VL-7B        & 0.70(0.018) & 0.81(0.026) & 0.79(0.041)    & 0.93(0.020) & 0.80(0.014) & 0.88(0.019) & 0.91(0.045)    & 0.94(0.014) \\
DeepSeek-VL-7B       & 0.46(0.019) & 0.60(0.033) & 0.42(0.047)    & 0.64(0.040) & 0.52(0.017) & 0.74(0.026) & 0.44(0.073)    & 0.66(0.030) \\
LLaVA-1.6-Mistral-7B & 0.46(0.020) & 0.70(0.031) & 0.39(0.048)    & 0.63(0.039) & 0.53(0.017) & 0.74(0.024) & 0.42(0.077)    & 0.60(0.030) \\
LLaVA-1.6-Vicuna-7B  & 0.41(0.020) & 0.57(0.034) & 0.39(0.047)    & 0.59(0.042) & 0.49(0.016) & 0.71(0.027) & 0.44(0.074)    & 0.56(0.031) \\
BLIP2-OPT-6.7B       & 0.19(0.016) & 0.22(0.028) & 0.20(0.040)    & 0.20(0.034) & 0.22(0.014) & 0.23(0.023) & 0.14(0.051)    & 0.24(0.027) \\\midrule
                     & \multicolumn{8}{c}{Domain-specific - Medical VL Model}                                                              \\\cmidrule{2-9}
LLaVA-Med            & 0.29(0.018) & 0.30(0.031) & 0.25(0.042)    & 0.30(0.039) & 0.26(0.015) & 0.31(0.027) & 0.14(0.053)    & 0.21(0.026) \\
Med-flamingo         & 0.27(0.017) & 0.27(0.029) & 0.25(0.042)    & 0.25(0.037) & 0.26(0.014) & 0.27(0.025) & 0.33(0.072)    & 0.25(0.027) \\\midrule
                     & \multicolumn{8}{c}{Domain-specific - Pathology VL Model}                                                            \\\cmidrule{2-9}
PA-LLaVA             & 0.23(0.016) & 0.25(0.027) & 0.28(0.044)    & 0.25(0.037) & 0.00(0.000) & 0.00(0.000) & 0.00(0.000)    & 0.00(0.000) \\
Quilt-LLaVA          & 0.34(0.018) & 0.41(0.032) & 0.36(0.047)    & 0.39(0.042) & 0.39(0.016) & 0.38(0.027) & 0.49(0.075)    & 0.41(0.031) \\
PathGen-LLaVA        & 0.76(0.016) & 0.85(0.024) & 0.75(0.041)    & 0.78(0.036) & 0.81(0.013) & 0.88(0.018) & 0.74(0.066)    & 0.60(0.032) \\
Ours - PBS-VL        & 0.94(0.009) & 0.96(0.012) & 0.98(0.013)    & 0.99(0.007) & 0.94(0.008) & 0.97(0.010) & 0.93(0.037)    & 0.99(0.007) \\\bottomrule
\end{tabular}%
}
\end{table*}

\begin{table*}[ht]
\centering
\caption{The partial match rate of cell-level \textbf{Fill-in-the-blank} questions for benchmarked models breaks down by task types.}
\label{tab:supp_result_fb}
\resizebox{0.9\textwidth}{!}{%
\begin{tabular}{l|cccc|cccc}
\toprule
                     & \multicolumn{4}{|c|}{In-domain}                            & \multicolumn{4}{c}{Out of Domain}                        \\\cmidrule{2-9}
                     & Morphology  & Abnormality & Cell subtyping & Knowledge   & Morphology  & Abnormality & Cell subtyping & Knowledge   \\\midrule
                     & \multicolumn{8}{c}{Proprietary VL Models}                                                                           \\\cmidrule{2-9}
GPT-4o               & 0.40(0.023) & 0.62(0.048) & 0.57(0.028)    & 0.34(0.074) & 0.23(0.016) & 0.24(0.038) & 0.51(0.055)    & 0.40(0.037) \\
Gemini-2.5-pro       & 0.36(0.022) & 0.50(0.049) & 0.47(0.027)    & 0.34(0.072) & 0.17(0.015) & 0.22(0.038) & 0.36(0.052)    & 0.44(0.037) \\
Claude-4.5-Sonnet    & 0.50(0.024) & 0.62(0.047) & 0.58(0.027)    & 0.54(0.078) & 0.22(0.016) & 0.23(0.040) & 0.49(0.053)    & 0.45(0.037) \\\midrule
                     & \multicolumn{8}{c}{Open-source General-purpose VL Model}                                                            \\\cmidrule{2-9}
InternVL-8B          & 0.34(0.022) & 0.46(0.050) & 0.44(0.029)    & 0.24(0.065) & 0.19(0.015) & 0.16(0.033) & 0.27(0.047)    & 0.17(0.028) \\
Idefics3-8B          & 0.15(0.017) & 0.36(0.047) & 0.16(0.021)    & 0.10(0.047) & 0.11(0.012) & 0.07(0.023) & 0.06(0.026)    & 0.18(0.029) \\
Qwen2-VL-7B          & 0.24(0.020) & 0.47(0.049) & 0.24(0.024)    & 0.20(0.061) & 0.12(0.013) & 0.20(0.036) & 0.09(0.031)    & 0.09(0.021) \\
Qwen2.5-VL-7B        & 0.14(0.016) & 0.39(0.048) & 0.16(0.020)    & 0.07(0.040) & 0.10(0.011) & 0.15(0.033) & 0.15(0.038)    & 0.06(0.017) \\
DeepSeek-VL-7B       & 0.12(0.015) & 0.40(0.048) & 0.12(0.019)    & 0.02(0.024) & 0.06(0.009) & 0.13(0.031) & 0.12(0.035)    & 0.02(0.011) \\
LLaVA-1.6-Mistral-7B & 0.12(0.015) & 0.27(0.045) & 0.03(0.009)    & 0.10(0.048) & 0.08(0.010) & 0.05(0.020) & 0.08(0.029)    & 0.18(0.029) \\
LLaVA-1.6-Vicuna-7B  & 0.14(0.016) & 0.23(0.040) & 0.02(0.008)    & 0.02(0.025) & 0.08(0.010) & 0.04(0.018) & 0.05(0.023)    & 0.15(0.027) \\
BLIP2-OPT-6.7B       & 0.18(0.018) & 0.39(0.048) & 0.21(0.022)    & 0.17(0.057) & 0.21(0.015) & 0.27(0.039) & 0.20(0.042)    & 0.17(0.027) \\\midrule
                     & \multicolumn{8}{c}{Domain-specific - Medical VL Model}                                                              \\\cmidrule{2-9}
LLaVA-Med            & 0.07(0.012) & 0.42(0.047) & 0.04(0.011)    & 0.00(0.000) & 0.06(0.009) & 0.15(0.032) & 0.06(0.026)    & 0.03(0.013) \\
Med-flamingo         & 0.03(0.009) & 0.26(0.044) & 0.04(0.011)    & 0.00(0.000) & 0.03(0.006) & 0.13(0.030) & 0.09(0.033)    & 0.01(0.006) \\\midrule
                     & \multicolumn{8}{c}{Domain-specific - Pathology VL Model}                                                            \\\cmidrule{2-9}
PA-LLaVA             & 0.00(0.000) & 0.00(0.000) & 0.00(0.000)    & 0.00(0.000) & 0.00(0.000) & 0.00(0.000) & 0.00(0.000)    & 0.00(0.000) \\
Quilt-LLaVA          & 0.15(0.016) & 0.37(0.047) & 0.11(0.017)    & 0.15(0.054) & 0.10(0.011) & 0.16(0.034) & 0.13(0.036)    & 0.08(0.020) \\
PathGen-LLaVA        & 0.34(0.023) & 0.41(0.047) & 0.25(0.024)    & 0.12(0.049) & 0.21(0.016) & 0.19(0.033) & 0.26(0.046)    & 0.11(0.024) \\
Ours - PBS-VL        & 0.41(0.023) & 0.59(0.048) & 0.74(0.024)    & 0.41(0.078) & 0.32(0.017) & 0.24(0.036) & 0.45(0.053)    & 0.47(0.036) \\\bottomrule
\end{tabular}%
}
\end{table*}

\begin{table*}[ht]
\centering
\caption{The semantic similarity of cell-level \textbf{Open-ended} questions for benchmarked models breaks down by task types.}
\label{tab:supp_result_open}
\resizebox{0.9\textwidth}{!}{%
\begin{tabular}{l|cccc|cccc}
\toprule
                     & \multicolumn{4}{|c|}{In-domain}                            & \multicolumn{4}{c}{Out of Domain}                        \\\cmidrule{2-9}
                     & Morphology  & Abnormality & Cell subtyping & Knowledge   & Morphology  & Abnormality & Cell subtyping & Knowledge   \\\midrule
                     & \multicolumn{8}{c}{Proprietary VL Models}                                                                           \\\cmidrule{2-9}
GPT-4o               & 0.70(0.013)  & 0.62(0.006)  & 0.77(0.010)    & 0.63(0.015)  & 0.74(0.010)  & 0.62(0.004) & 0.78(0.020)    & 0.68(0.009) \\
Gemini-2.5-pro       & 0.66(0.012)  & 0.58(0.007)  & 0.74(0.012)    & 0.54(0.015)  & 0.68(0.011)  & 0.54(0.005) & 0.73(0.019)    & 0.58(0.010) \\
Claude-4.5-Sonnet    & 0.58(0.022)  & 0.47(0.012)  & 0.69(0.023)    & 0.45(0.029)  & 0.66(0.009)  & 0.58(0.004) & 0.70(0.016)    & 0.62(0.008) \\\midrule
                     & \multicolumn{8}{c}{Open-source General-purpose VL Model}                                                            \\\cmidrule{2-9}
InternVL-8B          & 0.70(0.013)  & 0.59(0.008)  & 0.80(0.010)    & 0.62(0.017)  & 0.73(0.012)  & 0.59(0.005) & 0.78(0.021)    & 0.67(0.010) \\
Idefics3-8B          & 0.64(0.015)  & 0.52(0.009)  & 0.69(0.018)    & 0.56(0.018)  & 0.67(0.016)  & 0.52(0.006) & 0.68(0.028)    & 0.61(0.012) \\
Qwen2-VL-7B          & 0.69(0.012)  & 0.60(0.006)  & 0.75(0.010)    & 0.58(0.016)  & 0.74(0.011)  & 0.57(0.006) & 0.77(0.019)    & 0.63(0.011) \\
Qwen2.5-VL-7B        & 0.62(0.012)  & 0.57(0.006)  & 0.72(0.009)    & 0.57(0.014)  & 0.66(0.010)  & 0.55(0.005) & 0.72(0.016)    & 0.60(0.010) \\
DeepSeek-VL-7B       & 0.58(0.017)  & 0.50(0.009)  & 0.67(0.018)    & 0.50(0.017)  & 0.61(0.017)  & 0.49(0.006) & 0.68(0.013)    & 0.51(0.013) \\
LLaVA-1.6-Mistral-7B & 0.61(0.012)  & 0.53(0.007)  & 0.68(0.012)    & 0.56(0.015)  & 0.64(0.011)  & 0.53(0.005) & 0.71(0.024)    & 0.58(0.010) \\
LLaVA-1.6-Vicuna-7B  & 0.60(0.013)  & 0.51(0.006)  & 0.66(0.010)    & 0.54(0.013)  & 0.64(0.010)  & 0.51(0.005) & 0.67(0.024)    & 0.56(0.010) \\
BLIP2-OPT-6.7B       & 0.45(0.019)  & 0.38(0.008)  & 0.52(0.027)    & 0.39(0.017)  & 0.46(0.020)  & 0.34(0.007) & 0.45(0.043)    & 0.35(0.016) \\\midrule
                     & \multicolumn{8}{c}{Domain-specific - Medical VL Model}                                                              \\\cmidrule{2-9}
LLaVA-Med            & 0.62(0.013)  & 0.53(0.007)  & 0.69(0.012)    & 0.54(0.016)  & 0.67(0.012)  & 0.52(0.006) & 0.72(0.025)    & 0.59(0.012) \\
Med-flamingo         & 0.44(0.017)  & 0.43(0.008)  & 0.54(0.024)    & 0.36(0.016)  & 0.45(0.017)  & 0.38(0.006) & 0.42(0.042)    & 0.36(0.011) \\\midrule
                     & \multicolumn{8}{c}{Domain-specific - Pathology VL Model}                                                            \\\cmidrule{2-9}
PA-LLaVA             & -0.04(0.004) & -0.06(0.002) & -0.05(0.007)   & -0.07(0.005) & -0.00(0.003) & 0.01(0.001) & -0.03(0.004)   & 0.00(0.003) \\
Quilt-LLaVA          & 0.68(0.014)  & 0.57(0.008)  & 0.74(0.011)    & 0.55(0.017)  & 0.71(0.012)  & 0.55(0.006) & 0.75(0.025)    & 0.59(0.011) \\
PathGen-LLaVA        & 0.69(0.013)  & 0.61(0.007)  & 0.80(0.010)    & 0.58(0.017)  & 0.70(0.012)  & 0.56(0.006) & 0.78(0.020)    & 0.61(0.012) \\
Ours - PBS-VL        & 0.75(0.013)  & 0.70(0.007)  & 0.82(0.014)    & 0.72(0.016)  & 0.79(0.012)  & 0.68(0.005) & 0.80(0.022)    & 0.72(0.011) \\\bottomrule
\end{tabular}%
}
\end{table*}

\begin{table*}[ht]
\centering
\caption{The performance of slide-level QAs for benchmarked models breaks down by task types.}
\label{tab:supp_result_slide}
\resizebox{0.9\textwidth}{!}{%
\begin{tabular}{cclcccccc}
\toprule
\multicolumn{1}{l}{}               & \multicolumn{1}{l}{}  &          & GPT-4o      & Gemini-2.5-pro & Claude-4.5  & HistoGPT    & SlideChat   & PBS-VL-Slide \\\midrule
\multirow{7}{*}{Morphology}        & T/F                   & Accuracy & 0.61(0.087) & 0.52(0.089)    & 0.74(0.081) & 0.19(0.069) & 0.77(0.074) & 0.94(0.045)  \\\cmidrule{2-9}
                                   & MCQ                   & Accuracy & 0.78(0.094) & 0.67(0.109)    & 0.28(0.104) & 0.28(0.102) & 0.83(0.088) & 0.94(0.056)  \\\cmidrule{2-9}
                                   & \multirow{2}{*}{FB}   & EMatch   & 0.00(0.000) & 0.00(0.000)    & 0.00(0.000) & 0.00(0.000) & 0.00(0.000) & 0.25(0.098)  \\
                                   &                       & PMatch   & 0.00(0.000) & 0.05(0.049)    & 0.15(0.078) & 0.00(0.000) & 0.00(0.000) & 0.30(0.101)  \\\cmidrule{2-9}
                                   & \multirow{3}{*}{Open} & BLEU-1   & 0.10(0.000) & 0.02(0.000)    & 0.08(0.000) & 0.01(0.000) & 0.05(0.000) & 0.28(0.000)  \\
                                   &                       & ROGUE-L  & 0.16(0.000) & 0.08(0.000)    & 0.13(0.000) & 0.02(0.000) & 0.11(0.000) & 0.40(0.000)  \\
                                   &                       & Sim      & 0.82(0.000) & 0.74(0.000)    & 0.76(0.000) & 0.20(0.000) & 0.75(0.000) & 0.74(0.000)  \\\midrule
\multirow{6}{*}{Abnormality}       & MCQ                   & Accuracy & 0.83(0.149) & 0.67(0.193)    & 0.50(0.193) & 0.00(0.000) & 0.67(0.198) & 0.83(0.150)  \\\cmidrule{2-9}
                                   & \multirow{2}{*}{FB}   & EMatch   & 0.00(0.000) & 0.00(0.000)    & 0.00(0.000) & 0.00(0.000) & 0.00(0.000) & 0.00(0.000)  \\
                                   &                       & PMatch   & 0.00(0.000) & 0.00(0.000)    & 0.00(0.000) & 0.00(0.000) & 0.00(0.000) & 0.00(0.000)  \\\cmidrule{2-9}
                                   & \multirow{3}{*}{Open} & BLEU-1   & 0.12(0.025) & 0.07(0.019)    & 0.12(0.052) & 0.02(0.008) & 0.24(0.042) & 0.39(0.029)  \\
                                   &                       & ROGUE-L  & 0.22(0.027) & 0.12(0.029)    & 0.18(0.050) & 0.03(0.012) & 0.23(0.032) & 0.44(0.044)  \\
                                   &                       & Sim      & 0.76(0.027) & 0.70(0.007)    & 0.68(0.028) & 0.13(0.046) & 0.62(0.074) & 0.76(0.021)  \\\midrule
\multirow{4}{*}{Cell differential} & T/F                   & Accuracy & 0.25(0.214) & 0.25(0.209)    & 0.25(0.218) & 0.25(0.215) & 0.75(0.217) & 0.25(0.208)  \\\cmidrule{2-9}
                                   & MCQ                   & Accuracy & 0.69(0.090) & 0.85(0.071)    & 0.38(0.097) & 0.23(0.081) & 0.19(0.080) & 0.77(0.081)  \\\cmidrule{2-9}
                                   & \multirow{2}{*}{FB}   & EMatch   & 1.00(0.000) & 0.00(0.000)    & 0.00(0.000) & 0.00(0.000) & 0.00(0.000) & 1.00(0.000)  \\
                                   &                       & PMatch   & 1.00(0.000) & 1.00(0.000)    & 1.00(0.000) & 0.00(0.000) & 0.00(0.000) & 1.00(0.000)  \\\midrule
\multirow{3}{*}{Knowledge}         & \multirow{3}{*}{Open} & BLEU-1   & 0.18(0.013) & 0.09(0.009)    & 0.17(0.021) & 0.03(0.004) & 0.16(0.022) & 0.37(0.032)  \\
                                   &                       & ROGUE-L  & 0.19(0.015) & 0.11(0.009)    & 0.17(0.020) & 0.04(0.006) & 0.19(0.019) & 0.35(0.045)  \\
                                   &                       & Sim      & 0.73(0.041) & 0.65(0.042)    & 0.68(0.042) & 0.17(0.020) & 0.63(0.048) & 0.77(0.034)  \\\midrule
\multirow{4}{*}{Diagnosis}         & MCQ                   & Accuracy & 0.80(0.186) & 0.40(0.225)    & 0.20(0.180) & 0.40(0.210) & 0.80(0.175) & 1.00(0.000)  \\\cmidrule{2-9}
                                   & \multirow{3}{*}{Open} & BLEU-1   & 0.19(0.019) & 0.11(0.015)    & 0.15(0.015) & 0.04(0.005) & 0.15(0.033) & 0.34(0.036)  \\
                                   &                       & ROGUE-L  & 0.21(0.024) & 0.13(0.013)    & 0.19(0.020) & 0.05(0.005) & 0.23(0.019) & 0.43(0.037)  \\
                                   &                       & Sim      & 0.72(0.054) & 0.71(0.035)    & 0.65(0.044) & 0.14(0.026) & 0.58(0.060) & 0.76(0.040) \\\bottomrule
\end{tabular}%
}
\end{table*}

%% file: main.bib
@String(CVPR= {IEEE Conf. Comput. Vis. Pattern Recog.})

@String(ECCV= {Eur. Conf. Comput. Vis.})

@String(AAAI = {AAAI})

@String(CVPR  = {CVPR})

@String(ECCV  = {ECCV})

@misc{pbs_intro,
    author = {Cleveland Clinic},
    title = {Peripheral Blood Smear},
    howpublished = {\url{https://my.clevelandclinic.org/health/diagnostics/22742-peripheral-blood-smear-test}},
    note = {Accessed: 2025-10-14},
    year = {2022},
}

@inproceedings{He2020PathVQA,
  title={PathVQA: 30000+ Questions for Medical Visual Question Answering},
  author={He, Xuehai and Zhang, Yichen and Mou, Luntian and Xing, Eric and Xie, Pengtao},
  booktitle={AAAI Workshop on Health Intelligence},
  year={2020},
  url={https://arxiv.org/abs/2003.10286}
}

@article{Ikezogwo2023Quilt1M,
  title={{Quilt-1M}: One Million Image-Text Pairs for Histopathology},
  author={Ikezogwo, Wisdom O. and Seyfioglu, Mehmet S. and Ghezloo, Fatemeh and Geva, Dylan S. C. and Mohammed, Fatwir S. and Anand, Pavan K. and Krishna, Ranjay and Shapiro, Linda},
  journal={arXiv:2306.11207},
  year={2023},
  url={https://arxiv.org/abs/2306.11207}
}

@article{Sun2024PathMMU,
  title={{PathMMU}: A Massive Multimodal Expert-Level Benchmark for Understanding and Reasoning in Pathology},
  author={Sun, Yuxuan and Wu, Hao and Zhu, Chenglu and Zheng, Sunyi and others},
  journal={arXiv:2401.16355},
  year={2024},
  url={https://arxiv.org/abs/2401.16355}
}

@inproceedings{Chen2024WSIVQA,
  title={{WSI-VQA}: Interpreting Whole Slide Images by Generative Visual Question Answering},
  author={Chen, Pingyi and Zhu, Chenglu and Zheng, Sunyi and Li, Honglin and Yang, Lin},
  booktitle={ECCV},
  year={2024},
  url={https://www.ecva.net/papers/eccv_2024/papers_ECCV/papers/05355.pdf}
}

@inproceedings{Chen2025SlideChat,
  title={SlideChat: A Large Vision-Language Assistant for Whole-Slide Pathology Image Understanding},
  author={Chen, Ying and Wang, Guoan and Ji, Yuanfeng and Li, Yanjun and Ye, Jin and others},
  booktitle={CVPR},
  year={2025},
  url={https://openaccess.thecvf.com/content/CVPR2025/papers/Chen_SlideChat_A_Large_Vision-Language_Assistant_for_Whole-Slide_Pathology_Image_Understanding_CVPR_2025_paper.pdf}
}

@article{Liang2025WSILLAVA,
  title={{WSI-LLaVA}: A Multimodal Large Language Model for Whole Slide Image},
  author={Liang, Yuci and Lyu, Xinheng and Ding, Meidan and Chen, Wenting and Zhang, Jipeng and others},
  journal={arXiv:2412.02141},
  year={2025},
  url={https://arxiv.org/abs/2412.02141}
}

@article{Kefeli2024TCGAReports,
  title={{TCGA-Reports}: A Machine-Readable Pathology Reports Dataset for The Cancer Genome Atlas},
  author={Kefeli, J and others},
  journal={Scientific Data},
  year={2024},
  url={https://www.ncbi.nlm.nih.gov/pmc/articles/PMC10935496/}
}

@article{Chen2023WsiCaption,
  title={{WsiCaption}: Multiple Instance Generation of Pathology Reports for Gigapixel Whole-Slide Images},
  author={Chen, Pingyi and Li, Honglin and Zhu, Chenglu and Zheng, Sunyi and Shui, Zhongyi and Yang, Lin},
  journal={arXiv:2311.16480},
  year={2023},
  url={https://arxiv.org/abs/2311.16480}
}

@article{Chen2024UNI,
  title        = {Towards a general-purpose foundation model for computational pathology},
  author       = {Chen, Richard J. and others},
  journal      = {Nature Medicine},
  year         = {2024},
  doi          = {10.1038/s41591-024-03141-0},
  url          = {https://www.nature.com/articles/s41591-024-03141-0}
}

@article{Vorontsov2024Virchow,
  title        = {A foundation model for clinical-grade computational pathology},
  author       = {Vorontsov, Eugene and others},
  journal      = {Nature Medicine},
  year         = {2024},
  doi          = {10.1038/s41591-024-03141-0},
  url          = {https://www.nature.com/articles/s41591-024-03141-0}
}

@misc{Zimmermann2024Virchow2,
  title        = {virchow2: scaling self-supervised mixed magnification pretraining for computational pathology},
  author       = {Zimmermann, Eric and others},
  year         = {2024},
  eprint       = {2408.00738},
  archivePrefix= {arXiv},
  primaryClass = {cs.CV},
  url          = {https://arxiv.org/abs/2408.00738}
}

@article{Xu2024ProvGigaPath,
  title        = {A whole-slide foundation model for digital pathology from real-world data},
  author       = {Xu, Han and others},
  journal      = {Nature},
  year         = {2024},
  doi          = {10.1038/s41586-024-07441-w},
  url          = {https://www.nature.com/articles/s41586-024-07441-w}
}

@inproceedings{Koch2024DinoBloom,
  title        = {DinoBloom: A Foundation Model for Generalizable Cell Embeddings in Hematology},
  author       = {Koch, Vincent and others},
  booktitle    = {MICCAI},
  year         = {2024},
  doi          = {10.1007/978-3-031-72390-2_49},
  url          = {https://papers.miccai.org/miccai-2024/paper/3584_paper.pdf}
}

@misc{Bioptimus2025HOptimus1,
  title        = {H-optimus-1: Foundation Models for Histology},
  author       = {{Bioptimus}},
  year         = {2025},
  url          = {https://www.bioptimus.com/h-optimus-1},
  note         = {Company/model announcement}
}

@misc{Bioptimus2024HOptimus0,
  title        = {H-optimus-0: the world's largest open-source AI foundation model for pathology},
  author       = {{Bioptimus}},
  year         = {2024},
  url          = {https://www.bioptimus.com/news/bioptimus-launches-h-optimus-0-the-worlds-largest-open-source-ai-foundation-model-for-pathology}
}

@article{Huang2023PLIP,
  title        = {A visual–language foundation model for pathology image analysis},
  author       = {Huang, Ziyi and others},
  journal      = {Nature Medicine},
  year         = {2023},
  doi          = {10.1038/s41591-023-02469-0},
  url          = {https://pubmed.ncbi.nlm.nih.gov/37592105/}
}

@article{Lu2024CONCH,
  title        = {A visual-language foundation model for computational pathology},
  author       = {Lu, Ming Y. and others},
  journal      = {Nature Medicine},
  year         = {2024},
  doi          = {10.1038/s41591-024-02856-4},
  url          = {https://www.nature.com/articles/s41591-024-02856-4}
}

@article{Xiang2025MUSK,
  title        = {A vision–language foundation model for precision oncology},
  author       = {Xiang, Jinxi and others},
  journal      = {Nature Cancer},
  year         = {2025},
  doi          = {10.1038/s43018-025-00923-4},
  url          = {https://www.nature.com/articles/s43018-025-00923-4}
}

@misc{Shaikovski2024PRISM,
  title        = {{PRISM}: A Multi-Modal Generative Foundation Model for Slide-Level Histopathology},
  author       = {Shaikovski, George and others},
  year         = {2024},
  eprint       = {2405.10254},
  archivePrefix= {arXiv},
  primaryClass = {cs.CV},
  url          = {https://arxiv.org/abs/2405.10254}
}

@misc{CHIEF2024,
  title        = {{CHIEF}: Clinical Histopathology Imaging Evaluation Foundation Model},
  author       = {{HMS DBMI}},
  year         = {2024},
  url          = {https://github.com/hms-dbmi/CHIEF},
  note         = {Code repository and model card}
}

@article{TITAN2025,
  title        = {{TITAN}: A multimodal whole-slide foundation model for pathology},
  author       = {Lu, Ming Y. and others},
  journal      = {Nature Medicine},
  year         = {2025},
  doi          = {10.1038/s41591-025-03982-3},
  url          = {https://www.nature.com/articles/s41591-025-03982-3}
}

@article{Tran2025HistoGPT,
  title        = {Generating dermatopathology reports from gigapixel whole-slide images},
  author       = {Tran, Minh and others},
  journal      = {Nature Communications},
  year         = {2025},
  doi          = {10.1038/s41467-025-60014-x},
  url          = {https://www.nature.com/articles/s41467-025-60014-x}
}

@misc{Seyfioglu2023QuiltLLaVA,
  title        = {Quilt-{LLaVA}: Visual Instruction Tuning by Extracting Localized Narratives from Open-Source Histopathology Videos},
  author       = {Seyfioglu, Mehmet Saygin and others},
  year         = {2023},
  eprint       = {2312.04746},
  archivePrefix= {arXiv},
  primaryClass = {cs.CV},
  url          = {https://arxiv.org/abs/2312.04746}
}

@misc{Sun2024PathGen16M,
  title        = {PathGen-1.6M: 1.6 Million Pathology Image-text Pairs Generation through Multi-agent Collaboration},
  author       = {Sun, Yuxuan and others},
  year         = {2024},
  eprint       = {2407.00203},
  archivePrefix= {arXiv},
  primaryClass = {cs.CV},
  url          = {https://arxiv.org/abs/2407.00203}
}

@misc{Dai2024PALLaVA,
  title        = {PA-{LLaVA}: A Large Language-Vision Assistant for Human Pathology Image Understanding},
  author       = {Dai, Dawei and others},
  year         = {2024},
  eprint       = {2408.09530},
  archivePrefix= {arXiv},
  primaryClass = {cs.CV},
  url          = {https://arxiv.org/abs/2408.09530}
}

@article{Sun2024PathAsst,
	title = {{PathAsst}: {A} {Generative} {Foundation} {AI} {Assistant} towards {Artificial} {General} {Intelligence} of {Pathology}},
	volume = {38},
	issn = {2374-3468, 2159-5399},
	shorttitle = {{PathAsst}},
	url = {https://ojs.aaai.org/index.php/AAAI/article/view/28308},
	doi = {10.1609/aaai.v38i5.28308},
	language = {en},
	number = {5},
	urldate = {2025-03-17},
	journal = {Proceedings of the AAAI Conference on Artificial Intelligence},
	author = {Sun, Yuxuan and Zhu, Chenglu and Zheng, Sunyi and Zhang, Kai and Sun, Lin and Shui, Zhongyi and Zhang, Yunlong and Li, Honglin and Yang, Lin},
	month = mar,
	year = {2024},
	pages = {5034--5042},
}

@inproceedings{Sun2025CpathOmni,
  title={Cpath-omni: A unified multimodal foundation model for patch and whole slide image analysis in computational pathology},
  author={Sun, Yuxuan and Si, Yixuan and Zhu, Chenglu and Gong, Xuan and Zhang, Kai and Chen, Pingyi and Zhang, Ye and Shui, Zhongyi and Lin, Tao and Yang, Lin},
  booktitle={Proceedings of the Computer Vision and Pattern Recognition Conference},
  pages={10360--10371},
  year={2025}
}

@article{De2023computational,
  title={Computational analysis of peripheral blood smears detects disease-associated cytomorphologies},
  author={de Almeida, Jos{\'e} Guilherme and Gudgin, Emma and Besser, Martin and Dunn, William G and Cooper, Jonathan and Haferlach, Torsten and Vassiliou, George S and Gerstung, Moritz},
  journal={Nature Communications},
  volume={14},
  number={1},
  pages={4378},
  year={2023},
  publisher={Nature Publishing Group UK London}
}

@article{Pachitariu2025CellposeSAM,
  title   = {Cellpose-SAM: Superhuman generalization for cellular segmentation},
  author  = {Pachitariu, Marius and Rariden, Michael and Stringer, Carsen},
  journal = {bioRxiv},
  year    = {2025},
  month   = {May},
  publisher = {Cold Spring Harbor Laboratory},
  doi     = {10.1101/2025.04.28.651001},
  url     = {https://www.biorxiv.org/content/10.1101/2025.04.28.651001v1}
}

@misc{Matek2019AMLLMU,
  author    = {Matek, Christian and Schwarz, Simone and Marr, Carsten and Spiekermann, Karsten},
  title     = {A Single-cell Morphological Dataset of Leukocytes from AML Patients and Non-malignant Controls},
  year      = {2019},
  publisher = {The Cancer Imaging Archive},
  doi       = {10.7937/TCIA.2019.36F5O9LD},
  url       = {https://doi.org/10.7937/tcia.2019.36f5o9ld},
  note      = {Data set}
}

@misc{Shenderov2020APLKaggle,
  author       = {Shenderov, Eugene},
  title        = {Acute Promyelocytic Leukemia (APL) Peripheral Blood Smear Images},
  year         = {2020},
  howpublished = {\url{https://www.kaggle.com/datasets/eugeneshenderov/acute-promyelocytic-leukemia-apl}},
  note         = {Dataset on Kaggle; accessed 2025-11-10}
}

@misc{Lisc2022WBCLisc,
    title = { WBC LISC Dataset },
    type = { Open Source Dataset },
    author = { WBCs },
    howpublished = { \url{ https://universe.roboflow.com/wbcs/wbc-lisc } },
    url = { https://universe.roboflow.com/wbcs/wbc-lisc },
    journal = { Roboflow Universe },
    publisher = { Roboflow },
    year = { 2022 },
    month = { may },
    note = { visited on 2025-11-10 },
}

@article{rezatofighi2011automatic,
  title={Automatic recognition of five types of white blood cells in peripheral blood},
  author={Rezatofighi, Seyed Hamid and Soltanian-Zadeh, Hamid},
  journal={Computerized Medical Imaging and Graphics},
  volume={35},
  number={4},
  pages={333--343},
  year={2011},
  publisher={Elsevier}
}

@inproceedings{
  wang2023evaluating,
  title={Evaluating Open-{QA} Evaluation},
  author={Cunxiang Wang and Sirui Cheng and Qipeng Guo and Yuanhao Yue and Bowen Ding and Zhikun Xu and Yidong Wang and Xiangkun Hu and Zheng Zhang and Yue Zhang},
  booktitle={Thirty-seventh Conference on Neural Information Processing Systems Datasets and Benchmarks Track},
  year={2023},
  url={https://openreview.net/forum?id=UErNpveP6R}
}

@inproceedings{lin2004rouge,
  title={Rouge: A package for automatic evaluation of summaries},
  author={Lin, Chin-Yew},
  booktitle={Text summarization branches out},
  pages={74--81},
  year={2004}
}

@inproceedings{papineni2002bleu,
  title={Bleu: a method for automatic evaluation of machine translation},
  author={Papineni, Kishore and Roukos, Salim and Ward, Todd and Zhu, Wei-Jing},
  booktitle={Proceedings of the 40th annual meeting of the Association for Computational Linguistics},
  pages={311--318},
  year={2002}
}

@inproceedings{reimers2019SentenceBert,
    title = "Sentence-BERT: Sentence Embeddings using Siamese BERT-Networks",
    author = "Reimers, Nils and Gurevych, Iryna",
    booktitle = "Proceedings of the 2019 Conference on Empirical Methods in Natural Language Processing",
    month = "11",
    year = "2019",
    publisher = "Association for Computational Linguistics",
    url = "https://arxiv.org/abs/1908.10084",
}

@inproceedings{li2023blip2,
  title={Blip-2: Bootstrapping language-image pre-training with frozen image encoders and large language models},
  author={Li, Junnan and Li, Dongxu and Savarese, Silvio and Hoi, Steven},
  booktitle={International conference on machine learning},
  pages={19730--19742},
  year={2023},
  organization={PMLR}
}

@inproceedings{jaegle2021perceiver,
  title={Perceiver: General perception with iterative attention},
  author={Jaegle, Andrew and Gimeno, Felix and Brock, Andy and Vinyals, Oriol and Zisserman, Andrew and Carreira, Joao},
  booktitle={International conference on machine learning},
  pages={4651--4664},
  year={2021},
  organization={PMLR}
}

@article{li2023LlavaMed,
  title={Llava-med: Training a large language-and-vision assistant for biomedicine in one day},
  author={Li, Chunyuan and Wong, Cliff and Zhang, Sheng and Usuyama, Naoto and Liu, Haotian and Yang, Jianwei and Naumann, Tristan and Poon, Hoifung and Gao, Jianfeng},
  journal={Advances in Neural Information Processing Systems},
  volume={36},
  pages={28541--28564},
  year={2023}
}

@inproceedings{moor2023MedFlamingo,
  title={Med-flamingo: a multimodal medical few-shot learner},
  author={Moor, Michael and Huang, Qian and Wu, Shirley and Yasunaga, Michihiro and Dalmia, Yash and Leskovec, Jure and Zakka, Cyril and Reis, Eduardo Pontes and Rajpurkar, Pranav},
  booktitle={Machine Learning for Health (ML4H)},
  pages={353--367},
  year={2023},
  organization={PMLR}
}

@article{wang2025internvl35,
  title={Internvl3. 5: Advancing open-source multimodal models in versatility, reasoning, and efficiency},
  author={Wang, Weiyun and Gao, Zhangwei and Gu, Lixin and Pu, Hengjun and Cui, Long and Wei, Xingguang and Liu, Zhaoyang and Jing, Linglin and Ye, Shenglong and Shao, Jie and others},
  journal={arXiv preprint arXiv:2508.18265},
  year={2025}
}

@article{laurenccon2024idefics3,
  title={Building and better understanding vision-language models: insights and future directions},
  author={Lauren{\c{c}}on, Hugo and Marafioti, Andr{\'e}s and Sanh, Victor and Tronchon, L{\'e}o},
  journal={arXiv preprint arXiv:2408.12637},
  year={2024}
}

@article{wang2024qwen2,
  title={Qwen2-vl: Enhancing vision-language model's perception of the world at any resolution},
  author={Wang, Peng and Bai, Shuai and Tan, Sinan and Wang, Shijie and Fan, Zhihao and Bai, Jinze and Chen, Keqin and Liu, Xuejing and Wang, Jialin and Ge, Wenbin and others},
  journal={arXiv preprint arXiv:2409.12191},
  year={2024}
}

@article{bai2025qwen25,
  title={Qwen2. 5-vl technical report},
  author={Bai, Shuai and Chen, Keqin and Liu, Xuejing and Wang, Jialin and Ge, Wenbin and Song, Sibo and Dang, Kai and Wang, Peng and Wang, Shijie and Tang, Jun and others},
  journal={arXiv preprint arXiv:2502.13923},
  year={2025}
}

@article{lu2024deepseekvl,
  title={Deepseek-vl: towards real-world vision-language understanding},
  author={Lu, Haoyu and Liu, Wen and Zhang, Bo and Wang, Bingxuan and Dong, Kai and Liu, Bo and Sun, Jingxiang and Ren, Tongzheng and Li, Zhuoshu and Yang, Hao and others},
  journal={arXiv preprint arXiv:2403.05525},
  year={2024}
}

@misc{liu2024llavanext,
  title={Llavanext: Improved reasoning, ocr, and world knowledge},
  author={Liu, Haotian and Li, Chunyuan and Li, Yuheng and Li, Bo and Zhang, Yuanhan and Shen, Sheng and Lee, Yong Jae},
  howpublished = { \url{ https://llava-vl.github.io/blog/2024-01-30-llava-next/ } },
  year={2024}
}

@misc{nci2025TCGA,
  author       = {{The Cancer Genome Atlas (TCGA) Research Network}},
  title        = {The Cancer Genome Atlas (TCGA) Program},
  year         = {2025},
  howpublished = {\url{https://www.cancer.gov/ccg/research/genome-sequencing/tcga}},
  note         = {National Cancer Institute; accessed 2025-11-13}
}

@article{hurst2024gpt4o,
  title={Gpt-4o system card},
  author={Hurst, Aaron and Lerer, Adam and Goucher, Adam P and Perelman, Adam and Ramesh, Aditya and Clark, Aidan and Ostrow, AJ and Welihinda, Akila and Hayes, Alan and Radford, Alec and others},
  journal={arXiv preprint arXiv:2410.21276},
  year={2024}
}

@misc{anthropic2025Claude,
  author       = {Anthropic},
  title        = {Claude Sonnet 4.5},
  year         = {2025},
  howpublished = {\url{https://www.anthropic.com/claude/sonnet}},
  note         = {Model page; accessed 2025-11-13}
}

@misc{google2025Gemini,
  author       = {{Google DeepMind}},
  title        = {Gemini 2.5 Pro},
  year         = {2025},
  howpublished = {\url{https://deepmind.google/models/gemini/pro/}},
  note         = {Model page; accessed 2025-11-13}
}

@inproceedings{huang2017densenet,
  title={Densely connected convolutional networks},
  author={Huang, Gao and Liu, Zhuang and Van Der Maaten, Laurens and Weinberger, Kilian Q},
  booktitle={Proceedings of the IEEE conference on computer vision and pattern recognition},
  pages={4700--4708},
  year={2017}
}
